\documentclass{article}

\usepackage{PRIMEarxiv}
\usepackage[utf8]{inputenc}
\usepackage[T1]{fontenc}
\usepackage{hyperref}
\hypersetup{hidelinks}
\usepackage{url}
\usepackage{booktabs}
\usepackage{amsmath}
\usepackage{amsfonts}
\usepackage{graphicx}
\usepackage{multirow}
\usepackage{tabularx}
\usepackage{xspace}
\usepackage{subcaption}
\usepackage{xcolor}
\usepackage{colortbl}
\usepackage{pifont}
\usepackage{microtype}
\usepackage{float}
\usepackage{algorithm}
\usepackage{algorithmic}
\usepackage{docmute}
\usepackage{adjustbox}
\graphicspath{{figures/}}

\providecommand{\Description}[1]{}
\providecommand{\affiliation}[1]{}
\providecommand{\institution}[1]{}
\providecommand{\country}[1]{}
\providecommand{\shortauthors}{}

\newcommand{\method}{SAFER-DEIM\xspace}
\newcommand{\srf}{SRF\xspace}
\newcommand{\rcer}{RCER\xspace}
\newcommand{\desc}{\mathbf{d}}

\newcommand{\eg}{e.g.,\xspace}

\newcommand{\RR}{\mathbb{R}}

\title{Bridging Multimodal Fusion and Expert Routing via Spectral Reliability Descriptors for Robust Object Detection}

\author{
  Yefeng Wu \\
  wuyefengflc@163.com
}

\begin{document}
\maketitle

%%
%% Abstract
%%
\begin{abstract}
RGB-infrared detectors typically discard the statistics generated during cross-modal fusion, leaving downstream modules unaware of whether the current interaction is reliable.
We propose to extract a parameter-free, 7-dimensional spectral reliability descriptor---summarizing band energy, amplitude ratio, phase consistency, and cross-modal correlation---and to reuse it beyond the fusion stage.
The descriptor drives both Spectral Reliability Fusion (SRF), which gates a spectral residual against a conservative spatial base, and Reliability-Conditioned Expert Routing (RCER), which combines the descriptor with pooled content to steer sparse post-fusion experts.
Under matched ablations, descriptor-aware gating improves mAP50 over content-only adaptive gating; a $2{\times}2$ factorial analysis further shows that descriptor-conditioned routing provides the larger marginal gain over expert architecture alone at near-equal parameter count.
Under six synthetic degradations on DroneVehicle, average retention rises to 95.0\%, versus 92.0\% for content-only MoE and 87.9\% for concatenation, with the largest gain under modality drop; the same model also improves mAP50 by +5.2/+5.3 on the natural day/night split.
These results suggest that preserving fusion-time reliability as an explicit signal benefits both adaptive fusion and post-fusion conditional computation.
\end{abstract}

\keywords{Multimodal Object Detection, Cross-Modal Fusion, Spectral Reliability, Mixture of Experts, RGB-Infrared Detection}

%% ============================================================
%% 1. INTRODUCTION
%% ============================================================
\section{Introduction}
\label{sec:intro}

Multimodal detectors usually preserve the output of fusion, but not a representation of whether that fusion was trustworthy.
In RGB-infrared detection, this omission matters because many failures arise less from semantic ambiguity than from low light, blur, noise, misalignment, or partial modality loss, all of which weaken cross-modal agreement~\cite{liu2022tardalm3fd,sun2022dronevehicle}.
Once RGB and infrared features have been mixed into a fused tensor, downstream modules must infer reliability indirectly from content alone, even though content complexity and fusion reliability are different factors.

Most RGB-infrared pipelines address this issue only inside the fusion block.
Spatial-domain approaches~\cite{chen2023ignet,zhao2024emma} learn attention maps or gates to reweight modality contributions, while SuperFusion~\cite{tang2022superfusion} emphasizes alignment-aware fusion; frequency-domain methods~\cite{wang2024fredft,zhu2025wavemamba,li2025alignfreenet} capture band-specific complementarity between RGB texture and infrared saliency.
Yet the statistics produced during fusion are usually consumed locally and then discarded.
This is especially limiting once conditional computation is introduced: routers then adapt from fused content even though the hardest cases often stem from \emph{agreement collapse} between modalities rather than intrinsically complex semantics.

We view this missing information as a state variable that multimodal fusion should expose explicitly.
The frequency domain is a natural place to estimate such a state because cross-modal interaction there reveals interpretable cues about band energy, modal dominance, phase consistency, and cross-modal correlation.
We summarize these cues as a parameter-free, 7-dimensional \emph{spectral reliability descriptor} extracted from the same Fourier interaction already used for fusion.
Under clean, well-aligned conditions, the descriptor shows strong phase coherence, balanced amplitudes, and high cross-modal correlation; under blur, noise, misalignment, or modality loss, it changes in distinct and predictable ways.

Based on this descriptor, we build \method.
\srf uses the descriptor to gate spectral mixing against a conservative spatial fallback, while \rcer reuses the same descriptor together with pooled fused content to route sparse post-fusion experts.
The key idea is therefore not simply to add experts, but to change what drives adaptation: explicit fusion-time reliability rather than fused content alone.
We instantiate this design inside a DEIM-based detector, but the reuse principle itself is orthogonal to the detector head.

We evaluate this hypothesis with matched studies designed to isolate where the gains come from.
Descriptor-aware gating improves over fixed interpolation and content-only adaptive gating.
A $2{\times}2$ factorial analysis separates the routing signal from expert-bank design and shows that the former contributes the larger marginal gain.
Under blur, noise, misalignment, and modality drop, the advantage widens as cross-modal agreement deteriorates, which is consistent with the view that preserving fusion-time reliability benefits both adaptive fusion and downstream conditional computation.

Our main contributions are:
\begin{itemize}
  \item We formulate \textbf{fusion reliability as a reusable state variable} and instantiate it as a shared spectral reliability descriptor that preserves fusion-time evidence for later decisions in multimodal detection.
  \item We design \textbf{\srf} and \textbf{\rcer}, which reuse the same parameter-free descriptor for fusion gating and post-fusion conditional computation.
  \item We provide \textbf{matched empirical evidence} via controlled ablations, a $2{\times}2$ factorial routing study, robustness tests, and diagnostics showing that the descriptor matters most when cross-modal agreement degrades.
\end{itemize}

%% ============================================================
%% 2. RELATED WORK
%% ============================================================
\section{Related Work}
\label{sec:related}

\subsection{RGB-Infrared Fusion for Detection}

RGB-infrared fusion methods range from simple feature concatenation to structured interaction modules.
TarDAL~\cite{liu2022tardalm3fd} jointly optimizes image fusion and detection, while CDDFuse~\cite{zhao2023cddfuse} decomposes modalities into base/detail representations.
IGNet~\cite{chen2023ignet} and EMMA~\cite{zhao2024emma} use attention to reweight modality contributions, while SuperFusion~\cite{tang2022superfusion} focuses on alignment-aware fusion/registration mechanisms.
For weakly aligned settings, OAFA~\cite{chen2024oafa} further introduces adaptive alignment for UAV detection.
Recent work also questions when multimodal fusion is insufficient or unreliable: M$^2$D-LIF~\cite{zhao2025m2dlif} explicitly studies mono-modality insufficient learning and fusion degradation, reinforcing the need for reliability-aware downstream decisions rather than fusion alone.
Other recent efforts enrich multimodal detection with semantic priors~\cite{samguided2025acmmm}, knowledge distillation~\cite{decomkd2025acmmm}, or state-space modeling~\cite{cgssf2025acmmm}.
Orthogonally, general missing-modality modeling methods tackle degraded or missing modalities through structural separation or representation prediction; for example, missing-modality prediction methods~\cite{zheng2024mdqf} allow inference to proceed when one modality is absent.
Generic uncertainty modeling has also been studied in computer vision~\cite{kendall2017uncertainties}, while RGB-infrared detection work uses uncertainty as an auxiliary cue for cross-modal learning~\cite{sun2022dronevehicle}.
Both strategies address reliability concerns but derive their signal from content representations or architectural separation, whereas we extract reliability from the spectral statistics of the fusion process itself---a complementary signal source that requires no additional learned predictor and no architectural decoupling.

Most methods, however, still treat fusion as a one-way feature-construction stage.
Even when reliability or modality insufficiency motivates the design, the resulting statistics are consumed locally within the fusion block and are \emph{not} preserved as an explicit variable that later modules---such as routing or conditional computation---can act upon.

\subsection{Frequency-Domain Fusion and Conditional Computation}

Fourier-based neural operations provide global receptive fields at low computational cost~\cite{rao2021gfnet,chi2020ffc} and have been adopted in various vision tasks.
For multimodal detection, FreDFT~\cite{wang2024fredft} applies a frequency-domain fusion transformer to RGB-infrared inputs, modeling cross-modal interactions entirely in the spectral domain.
WaveMamba~\cite{zhu2025wavemamba} combines multi-scale wavelet decomposition with Mamba architectures, achieving strong results on M3FD, DroneVehicle, and FLIR-Aligned at the cost of 69.1M parameters.
AlignFreeNet~\cite{li2025alignfreenet} addresses alignment-free lightweight fusion, tackling the practical challenge of imperfect sensor registration.
RSDet~\cite{chen2024rsdet} takes a complementary approach, removing spectral redundancy between modalities via frequency-domain analysis to improve detection with reduced model size.
Sparse mixture-of-experts (MoE) architectures~\cite{shazeer2017moe} route different inputs to specialized sub-networks, achieving capacity scaling without proportional compute increase.
In multimodal detection, however, such conditional modules usually select experts from the content of the fused features alone.
More broadly, conditional computation has been explored through token pruning~\cite{rao2021dynamicvit} and dynamic detection heads~\cite{dai2021dynamichead}, which adapt computation based on input complexity; these mechanisms condition on content representations, and combining them with explicit reliability signals remains unexplored.

We combine these two directions.
\srf uses frequency interaction to construct fused features while exposing a compact descriptor of scene reliability, and \rcer uses that same descriptor to guide routing.
Relative to prior frequency-domain detectors, we retain fusion-time spectral evidence instead of consuming it locally inside the fusion block.
Relative to content-only MoE, the novelty is not the existence of experts itself, but the routing signal: explicit fusion-time reliability rather than content alone.

%% ============================================================
%% 3. METHOD
%% ============================================================
\section{Method}
\label{sec:method}

\subsection{Overview}
\label{sec:overview}

\begin{figure*}[t]
  \centering
  \includegraphics[width=\textwidth]{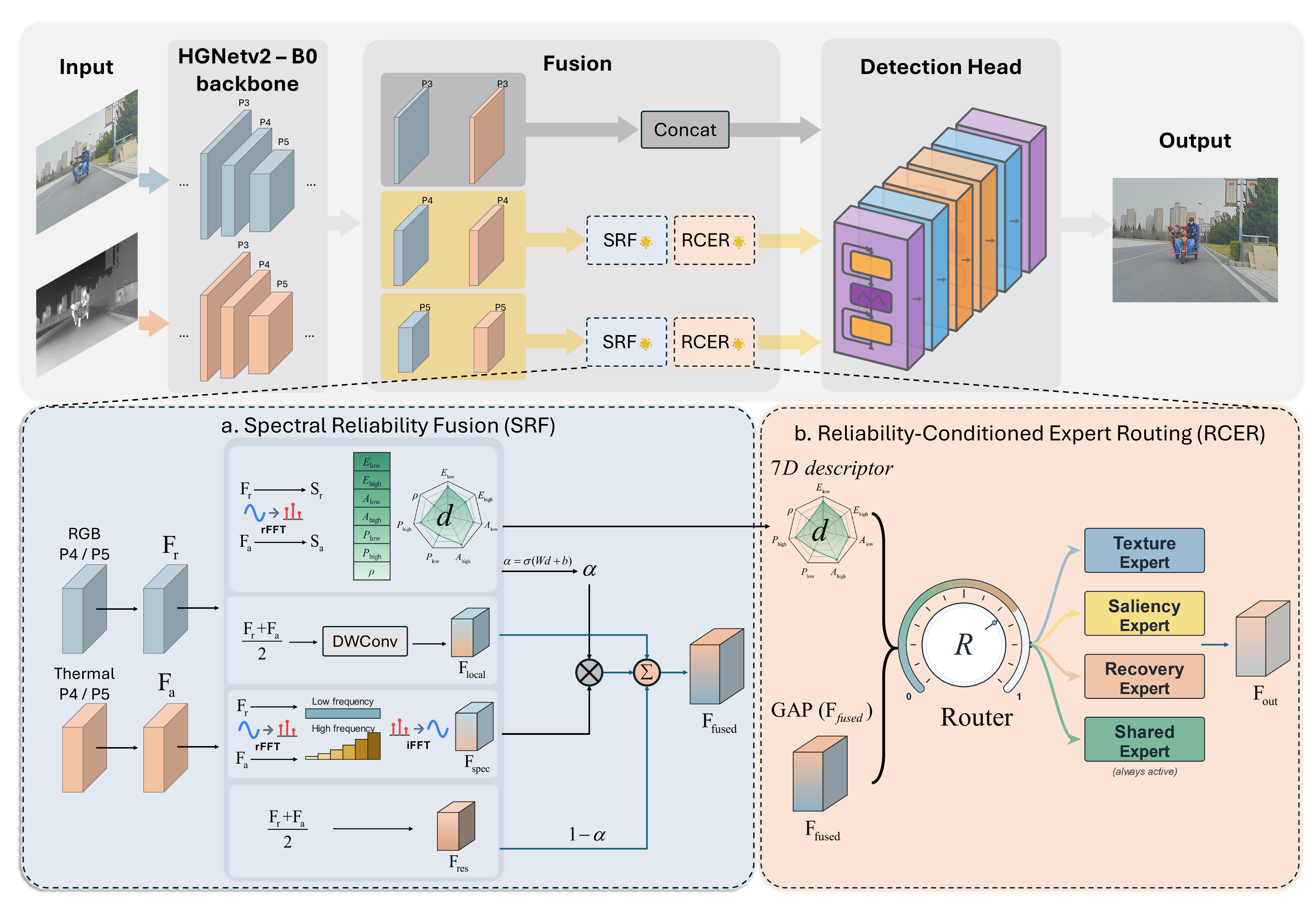}
  \caption{\textbf{\method overview.} Top: a dual-branch HGNetv2-B0 backbone extracts RGB and thermal features at $\{P_3, P_4, P_5\}$; $P_3$ uses plain concatenation, whereas $P_4$ and $P_5$ apply \srf followed by \rcer before the DEIM encoder and DFINETransformer decoder. (a) \srf mixes local, spectral, and residual branches to produce the fused feature $\mathbf{F}_\mathrm{fused}$ together with the 7D spectral reliability descriptor $\desc$. (b) \rcer reuses $\desc$ together with $\mathrm{GAP}(\mathbf{F}_\mathrm{fused})$ to route specialized experts while keeping the shared expert always active.}
  \Description{Overall detector pipeline with RGB and thermal inputs, HGNetv2-B0 backbone, P3 concatenation, SRF and RCER at P4 and P5, DEIM and DFINE detection head, and detailed panels showing descriptor generation in SRF and expert routing in RCER.}
  \label{fig:framework}
\end{figure*}

Figure~\ref{fig:framework} presents the overall architecture of \method.
A dual-branch HGNetv2-B0 backbone extracts features $\{P_3, P_4, P_5\}$ from RGB and thermal inputs.
At $P_4$ and $P_5$, \srf replaces plain concatenation and returns both fused features and a descriptor $\desc \in \RR^7$.
The same descriptor is immediately reused by \rcer at the same scales.
At $P_3$, we retain concatenation to avoid placing FFT and sparse-routing overhead on the highest-resolution map.
The resulting multi-scale features are then processed by the DEIM~\cite{ouyang2025deim} encoder and the DFINETransformer~\cite{zhang2022dfine} decoder.
The method follows a \emph{shared signal reuse} design: the descriptor is computed once during fusion and remains available for later conditional computation.
Panels~(a) and~(b) detail these two stages: \srf produces the descriptor during fusion, and \rcer combines it with $\mathrm{GAP}(\mathbf{F}_\mathrm{fused})$ to activate reliability-matched experts.

\subsection{Shared Spectral Reliability Descriptor}
\label{sec:descriptor}

At each active pyramid level, the method first aligns RGB and auxiliary features to a shared channel dimension $C$:
\begin{equation}
  \mathbf{F}_r = \text{BN}(\text{Conv}_{1\times1}(\mathbf{F}_\text{rgb})),\quad
  \mathbf{F}_a = \text{BN}(\text{Conv}_{1\times1}(\mathbf{F}_\text{aux})).
\end{equation}
We compute the 2D real FFT of both aligned tensors in FP32 for numerical stability under AMP:
\begin{equation}
  \mathbf{S}_r = \text{rFFT2D}(\mathbf{F}_r),\quad
  \mathbf{S}_a = \text{rFFT2D}(\mathbf{F}_a).
\end{equation}
The spectrum is divided by a binary radial mask with cutoff $r_0 = r_\text{max}\tau$, where we set $\tau=0.25$ (ablated in Section~\ref{sec:hyper}):
\begin{equation}
  \mathbf{S}_m^k = \mathbf{S}_m \odot \mathbf{M}_k,\quad
  m \in \{r,a\},\; k \in \{\text{low},\text{high}\}.
\end{equation}
We keep a binary low/high split rather than finer partitions because the design targets the main operational contrast in RGB-infrared detection: low-band structural/saliency agreement versus high-band texture reliability.

The descriptor is extracted from the \emph{reference spectrum}
\begin{equation}
  \widetilde{\mathbf{S}} = \tfrac{1}{2}(\mathbf{S}_r + \mathbf{S}_a),
\end{equation}
This choice is deliberate: the descriptor should summarize scene reliability, not become a self-referential by-product of adaptive high-band weighting.

From $\widetilde{\mathbf{S}}$, $\mathbf{S}_r$, and $\mathbf{S}_a$, the descriptor module computes seven scalar statistics per sample without learnable parameters:
\begin{align}
  E_k &= \text{mean}\!\left(|\widetilde{\mathbf{S}}^k|^2\right), \\
  A_k &= \frac{\text{mean}(|\mathbf{S}_r^k|)}{\text{mean}(|\mathbf{S}_a^k|) + \epsilon}, \\
  P_k &= \text{mean}\!\left(
    \frac{\operatorname{Re}(\mathbf{S}_r^k \odot \overline{\mathbf{S}_a^k})}
         {|\mathbf{S}_r^k||\mathbf{S}_a^k| + \epsilon}
  \right), \\
  \rho &= \operatorname{corr}(|\mathbf{S}_r|, |\mathbf{S}_a|).
\end{align}
Here $E_k$ is band energy, $A_k$ is amplitude ratio, $P_k$ is phase consistency, and $\rho$ is the global Pearson correlation of the flattened amplitude spectra.
Throughout, $\epsilon = 10^{-5}$ is a small constant for numerical stability.
The final descriptor is
\begin{equation}
  \desc = [E_\text{low},\, E_\text{high},\, A_\text{low},\, A_\text{high},\, P_\text{low},\, P_\text{high},\, \rho] \in \RR^7.
  \label{eq:descriptor}
\end{equation}

This design has two practical properties.
First, the descriptor is parameter-free and easy to inspect.
Second, its components are directly related to the degradations that matter in multimodal detection: blur mainly suppresses $E_\text{high}$, strong noise reduces $P_\text{high}$, and misalignment or modality drop strongly depresses $\rho$.
The implementation keeps the raw amplitude ratio rather than a log-ratio so that modality dominance remains directional: values above 1 indicate RGB-dominant energy, while values below 1 indicate auxiliary-dominant energy.

\textbf{Gradient flow and normalization.}
Before the descriptor is consumed, we apply
\begin{equation}
  \widehat{\desc} = \operatorname{LN}(\operatorname{stopgrad}(\desc)).
  \label{eq:descriptor_norm}
\end{equation}
Stopping gradients keeps the descriptor a measurement of the current input pair instead of letting the detection loss distort its scene-condition semantics.
LayerNorm compensates for the heterogeneous channel scales before linear projection.
The Pearson correlation $\rho$ is computed by flattening both $|\mathbf{S}_r|$ and $|\mathbf{S}_a|$ over all spatial-frequency bins and channel dimensions into 1D vectors and then applying the standard Pearson formula; if either flattened amplitude vector has variance below $\epsilon$ (\eg complete modality zeroing), we set $\rho=0$.
We use the notation $P_k$ (roman capital) for per-band phase consistency and $\rho$ (Greek) for global cross-modal correlation to distinguish these two complementary statistics.

\subsection{Spectral Reliability Fusion (SRF)}
\label{sec:srf}

\srf consumes the aligned features and the descriptor to produce a fused representation as a local spatial base combined with a reliability-gated blend.

\textbf{Local spatial branch.}
A depthwise $3\times3$ convolution followed by batch normalization and SiLU preserves local discriminative information that purely spectral processing may miss:
\begin{equation}
  \mathbf{F}_\text{local} = \text{DWConv}_{3\times3}\!\left(\tfrac{1}{2}(\mathbf{F}_r + \mathbf{F}_a)\right).
\end{equation}

\textbf{Spectral interaction branch.}
For the low band we apply uniform averaging, because both modalities contribute stable structural cues:
\begin{equation}
  \mathbf{S}_\text{low} = \tfrac{1}{2}(\mathbf{S}_r^\text{low} + \mathbf{S}_a^\text{low}).
\end{equation}
For the high band we use amplitude-adaptive weighting:
\begin{equation}
  \mathbf{w} = \frac{|\mathbf{S}_r^\text{high}|}{|\mathbf{S}_r^\text{high}| + |\mathbf{S}_a^\text{high}| + \epsilon},
\end{equation}
\begin{equation}
  \mathbf{S}_\text{high} = \mathbf{w}\cdot\mathbf{S}_r^\text{high} + (1-\mathbf{w})\cdot\mathbf{S}_a^\text{high}.
\end{equation}
\begin{equation}
  \mathbf{F}_\text{spec} = \text{irFFT2D}(\mathbf{S}_\text{low} + \mathbf{S}_\text{high}).
\end{equation}

\textbf{Reliability-gated output.}
A single linear projection $\mathbf{W} \in \RR^{C \times 7}$, $\mathbf{b} \in \RR^C$ maps the normalized descriptor to per-channel gate values:
\begin{equation}
  \boldsymbol{\alpha} = \sigma(\mathbf{W}\widehat{\desc} + \mathbf{b}),\quad \boldsymbol{\alpha} \in [0,1]^C.
\end{equation}
The fused output is constructed as a local base plus a gated blend:
\begin{equation}
  \mathbf{F}_\text{fused}
  = \underbrace{\mathbf{F}_\text{local}}_{\text{spatial base}}
  + \underbrace{\boldsymbol{\alpha} \cdot \mathbf{F}_\text{spec}
  + (1 - \boldsymbol{\alpha}) \cdot \tfrac{1}{2}(\mathbf{F}_r + \mathbf{F}_a)}_{\text{reliability-gated blend}}.
  \label{eq:gate}
\end{equation}
The local base preserves spatial discriminability regardless of the descriptor.
The residual term blends the spectrally interacted features with the conservative modality average, controlled by $\boldsymbol{\alpha}$: when the descriptor indicates strong phase consistency and cross-modal agreement, $\boldsymbol{\alpha}$ emphasizes the spectral branch; under degradation, the residual reverts toward the modality average, limiting the influence of unreliable spectral mixing.
Because the gate sees \emph{only} the descriptor and not pooled semantic content, the spectral-versus-spatial trade-off is forced to depend on reliability statistics rather than class-specific appearance.
\srf outputs both $\mathbf{F}_\text{fused}$ and $\desc$.

\subsection{Reliability-Conditioned Expert Routing (RCER)}
\label{sec:rcer}

The fused features from \srf enter \rcer for scene-adaptive post-fusion processing.
The central difference from content-only MoE is the routing signal: \rcer uses explicit scene-condition information from $\desc$ in addition to pooled fused content.

\textbf{Router architecture.}
The router receives the concatenation of a global content summary and the normalized reliability descriptor:
\begin{equation}
  \mathbf{r} = [\text{GAP}(\mathbf{F}_\text{fused});\; \widehat{\desc}] \in \RR^{C+7},
\end{equation}
where GAP is global average pooling.
The router is a two-layer MLP with hidden size $h=\max((C+7)/8, 8)$, LayerNorm, and SiLU.
It produces logits over 3 task experts.
During training, Gaussian noise ($\sigma{=}1.0$) is added to the logits before softmax to encourage exploration~\cite{shazeer2017moe}.
Top-2 experts are selected and their weights renormalized.

\textbf{Expert specialization.}
We use three task experts and one always-active shared expert:

\noindent\emph{Texture Expert}: a $1\times1$ projection, two stacked depthwise $3\times3$ convolutions with GroupNorm and SiLU, and a $1\times1$ output projection.
This branch is intended for clean scenes in which high-frequency detail is trustworthy.

\noindent\emph{Saliency Expert}: a $1\times1$ projection, a depthwise $5\times5$ convolution, efficient channel attention (ECA)~\cite{wang2020eca}, and a $1\times1$ output projection.
This branch is intended for low-light or low-contrast scenes.

\noindent\emph{Recovery Expert}: a $1\times1$ projection, two depthwise $3\times3$ convolutions with a residual connection, and a $1\times1$ output projection.
This branch is intended for blur, noise, misalignment, or modality corruption.

\noindent\emph{Shared Expert}: A $1{\times}1$ convolution with BatchNorm and SiLU, always activated.
It provides a stable baseline path even when routing becomes highly concentrated.

The final output combines the shared and routed expert outputs:
\begin{equation}
  \mathbf{F}_\text{out} = \mathbf{F}_\text{shared} + \sum_{i \in \text{top-2}} w_i \cdot \text{Expert}_i(\mathbf{F}_\text{fused}).
\end{equation}

\begin{table}[t]
\caption{\textbf{Descriptor-driven design rationale.} Intended mapping from scene conditions to descriptor cues and downstream responses.}
\label{tab:rationale}
\centering
\small
\setlength{\tabcolsep}{5pt}
\renewcommand{\arraystretch}{1.05}
\begin{tabularx}{\textwidth}{@{}l >{\raggedright\arraybackslash}X >{\raggedright\arraybackslash}p{0.20\textwidth} >{\raggedright\arraybackslash}p{0.17\textwidth}@{}}
\toprule
Condition & Descriptor cue & SRF response & RCER bias \\
\midrule
Clean / daytime & high $P_\text{high}$, high $\rho$, balanced $A_\text{high}$ & trust spectral branch & Texture \\
Low-light / night & weaker RGB high-band ratio, stable low-band energy & moderate spectral use & Saliency \\
Blur / noise & reduced $E_\text{high}$ or $P_\text{high}$ & increase fallback & Recovery \\
Misalign / drop & low $P_\text{high}$ and low $\rho$ & suppress unreliable fusion & Recovery / shared \\
\bottomrule
\end{tabularx}
\end{table}

\subsection{Design Rationale and Complexity}

Table~\ref{tab:rationale} summarizes the intended mechanism.
The descriptor is extracted from a pre-gating reference spectrum, inserted only at $P_4$ and $P_5$, and consumed by texture, saliency, and recovery experts chosen to match the recurring operating modes of RGB-infrared detection.
The descriptor extractor itself is parameter-free; the added learnable overhead is limited to one descriptor-to-channel gate in each \srf block and one lightweight router in each \rcer block.
Accordingly, the full detector remains lightweight at 19.8M parameters and 17.2 GFLOPs, and a near-equal comparison to content-only MoE is meaningful (19.6M vs.\ 19.8M in Table~\ref{tab:ablation}).
Under FP16 inference on a single RTX 4090, the proposed components add 1.3\,ms within a 9.1\,ms forward pass, with descriptor extraction accounting for only 0.3\,ms; the detailed latency breakdown is reported in Supplementary Table~S1.

\subsection{Training Objective}
\label{sec:loss}

The model is trained with the DEIM detection loss plus the sparse-MoE routing regularizer:
\begin{equation}
  \mathcal{L} = \mathcal{L}_\text{det} + \lambda_\text{moe} \mathcal{L}_\text{moe},
  \label{eq:total_loss}
\end{equation}
where $\mathcal{L}_\text{det}$ comprises Varifocal Loss~\cite{zhang2021varifocal}, bounding-box regression, and local alignment losses following the DEIM formulation~\cite{ouyang2025deim}, and $\mathcal{L}_\text{moe}$ combines the standard load-balancing loss from sparse MoE training~\cite{shazeer2017moe} with the router z-loss used to stabilize large sparse models~\cite{zoph2022stmoe}.
We set $\lambda_\text{moe}=0.01$ across all experiments.

%% ============================================================
%% 4. EXPERIMENTS
%% ============================================================
\section{Experiments}
\label{sec:exp}

Table~\ref{tab:main} provides contextual benchmark results, after which we focus on the matched tests that directly evaluate the core hypothesis.

\begin{table*}[!t]
\caption{\textbf{Published results on three RGB-infrared benchmarks.} Numbers are transcribed from the cited papers under their reported settings. Because backbones, detector heads, and training recipes differ, this table is used for contextual comparison only.}
\label{tab:main}
\centering
\footnotesize
\renewcommand{\arraystretch}{1.05}
\begin{tabularx}{\textwidth}{@{}>{\raggedright\arraybackslash}X c c c@{}}
\toprule
\multicolumn{4}{l}{\textit{Panel A. M3FD (mAP50 / mAP).}} \\
\midrule
Method & Backbone & mAP50 / mAP & Params (M) \\
\midrule
TarDAL~\cite{liu2022tardalm3fd} & YOLOv5 & 80.5 / 54.1 & -- \\
DetFusion~\cite{zhao2023detfusion} & YOLOv5 & 80.8 / 53.8 & -- \\
IGNet~\cite{chen2023ignet} & YOLOv5 & 81.5 / 54.5 & -- \\
EMMA~\cite{zhao2024emma} & YOLOv5 & 82.9 / 55.4 & -- \\
MMFN~\cite{wang2024mmfn} & YOLOv5 & 86.2 / 57.4 & -- \\
TSJNet~\cite{wang2024tsjnet} & YOLOv7 & 86.0 / 58.9 & -- \\
FreDFT~\cite{wang2024fredft} & YOLOv5 & 88.4 / 59.7 & 152.6 \\
WaveMamba~\cite{zhu2025wavemamba} & YOLOv8 & \textbf{92.1 / 64.4} & 69.1 \\
\rowcolor{gray!10}
\method & HGNetv2-B0 & 90.3 / 62.1 & \textbf{19.8} \\
\bottomrule
\end{tabularx}

\vspace{0.5em}
\begin{tabularx}{\textwidth}{@{}>{\raggedright\arraybackslash}X c c c@{}}
\toprule
\multicolumn{4}{l}{\textit{Panel B. DroneVehicle HBB (mAP50 / mAP).}} \\
\midrule
Method & Backbone & mAP50 / mAP & Params (M) \\
\midrule
IV-YOLO~\cite{kim2024ivyolo} & YOLOv8 & 74.6 / 56.8 & -- \\
M2FNet~\cite{wang2024cma} & YOLOv8 & 76.8 / 50.4 & -- \\
DAAB-FFPN~\cite{chen2024daabffpn} & YOLOv8 & 75.2 / 56.3 & -- \\
WaveMamba~\cite{zhu2025wavemamba} & YOLOv8 & 79.8 / 60.5 & 69.1 \\
\rowcolor{gray!10}
\method & HGNetv2-B0 & \textbf{80.5 / 62.0} & \textbf{19.8} \\
\bottomrule
\end{tabularx}

\vspace{0.5em}
\begin{tabularx}{\textwidth}{@{}>{\raggedright\arraybackslash}X c c c@{}}
\toprule
\multicolumn{4}{l}{\textit{Panel C. FLIR-Aligned (mAP50 / mAP).}} \\
\midrule
Method & Backbone & mAP50 / mAP & Params (M) \\
\midrule
MFPT~\cite{zhou2022mfpt} & ResNet50 & 80.0 / 41.9 & 200.0 \\
CrossFormer~\cite{li2023crossformer} & YOLOv5 & 79.3 / 42.1 & 340.0 \\
ESSFN~\cite{wang2024essfn} & YOLOv8 & 80.8 / 42.3 & 80.2 \\
WaveMamba~\cite{zhu2025wavemamba} & YOLOv8 & \textbf{88.4 / 48.1} & 69.1 \\
\rowcolor{gray!10}
\method & HGNetv2-B0 & 85.8 / 46.5 & \textbf{19.8} \\
\bottomrule
\end{tabularx}
\end{table*}

\subsection{Experimental Setup}
\label{sec:setup}

\textbf{Datasets.}
We evaluate on three public RGB-infrared benchmarks:
(1)~\textbf{M3FD}~\cite{liu2022tardalm3fd}: 4,200 aligned image pairs across 6 classes (people, car, bus, motorcycle, lamp, truck) captured in driving scenarios with day, night, and challenging weather conditions.
(2)~\textbf{DroneVehicle}~\cite{sun2022dronevehicle}: 28,439 pairs across 5 vehicle classes (car, freight car, truck, bus, van) from UAV perspective, featuring substantial day-night variation and natural sensor misalignment in our evaluation.
DroneVehicle was originally annotated with oriented bounding boxes (OBB); following recent work that evaluates horizontal bounding boxes for computational efficiency, we convert all annotations to axis-aligned HBB format.
This protocol is applied consistently across all models in our controlled studies.
(3)~\textbf{FLIR-Aligned}: derived from the official Teledyne FLIR ADAS dataset~\cite{flir2018dataset}; following WaveMamba~\cite{zhu2025wavemamba}, we use the aligned 4-class subset in urban road scenes, with 4,129 training pairs and 1,013 test pairs after removing the dog category.
For all datasets, auxiliary modality inputs are single-channel thermal images converted via min-max normalization to $[0, 255]$ and resized to match the RGB spatial resolution.
RGB images are loaded as standard 3-channel tensors.

\textbf{Implementation details.}
All models use dual HGNetv2-B0 backbones, $640{\times}640$ input resolution, AdamW optimizer (backbone lr $4{\times}10^{-4}$, detection head lr $8{\times}10^{-4}$), total batch size 4 across 4 GPUs, and 160 training epochs with automatic mixed-precision (AMP) and exponential moving average (EMA, decay 0.9999).
Mosaic augmentation and photometric distortion are applied for the first 148 epochs and disabled for the final 12 epochs to stabilize convergence.
\srf and \rcer are inserted at $P_4$ and $P_5$ with $\tau{=}0.25$, 3 task experts, and top-$k{=}2$.
Unless noted otherwise, the paper reports matched studies as mean$\pm$std over 5 seeds from this configuration.
For the core ablation (Table~\ref{tab:ablation}), all configurations use the same seeds and data splits to ensure fair comparison.

\textbf{Metrics.}
We report mAP@0.5 (mAP50) and mAP@[0.5:0.95] (mAP) following the COCO evaluation protocol.
For robustness evaluation, we additionally report retention rate:
\[
\text{Retention} = \text{mAP50}_\text{degraded} \,/\, \text{mAP50}_\text{clean} \times 100\%.
\]

\textbf{Baselines.}
All contextual baseline numbers are transcribed from the corresponding cited papers.
Because these methods use different backbones, detector heads, and training recipes, Table~\ref{tab:main} is used only for contextual positioning.
The main evidence for the proposed idea comes from the matched ablations and robustness tests.
For the MoE comparators in Table~\ref{tab:ablation}, all variants share the same sparse training scaffold (top-$k{=}2$ sparsity, shared expert, and MoE regularization).
The ``uniform experts + content-only router'' baseline uses a homogeneous expert bank with a content-only adaptive router, while the remaining rows systematically vary the routing signal and expert architecture to enable factorial analysis.

Table~\ref{tab:main} is used only for contextual positioning because backbones, heads, and training recipes differ across papers.
Within that caveat, \method remains competitive at a much smaller model budget: it reaches 90.3/62.1 on M3FD, 80.5/62.0 on DroneVehicle, and 85.8/46.5 on FLIR-Aligned with 19.8M parameters.
We also verified the trend under a second backbone: replacing HGNetv2-B0 with ResNet-50 while keeping SRF and RCER unchanged raises M3FD mAP50 from 82.1 to 87.4 (+5.3), close to the +5.8 gain observed under HGNetv2-B0.
Supplementary Sec.~S1 summarizes the matched cross-backbone result in Table~S2 and the contextual efficiency plot in Figure~S1.

\begin{table*}[!t]
\caption{\textbf{Controlled ablations under the HGNetv2-B0 recipe.} Panel A compares SRF gate variants on M3FD. Panel B reports four near-equal-parameter MoE rows forming a $2{\times}2$ factorial study over the routing signal and expert architecture, together with one lower-parameter SRF-only reference row outside the factorial design. Mean$\pm$std is reported for the matched five-seed runs where applicable.}
\label{tab:ablation}
\centering
\small
\setlength{\tabcolsep}{5pt}
\renewcommand{\arraystretch}{1.08}
\begin{tabularx}{\textwidth}{@{}>{\raggedright\arraybackslash}X c c >{\raggedright\arraybackslash}p{0.24\textwidth}@{}}
\toprule
\multicolumn{4}{l}{\textit{Panel A. Fusion gate ablation on M3FD.}} \\
\midrule
Variant & M3FD mAP50 & $\Delta$ vs.\ baseline & Reading \\
\midrule
Baseline concat & 84.5$\pm$0.2 & -- & no spectral branch \\
Local base + modality avg ($\alpha{=}0$) & 85.8 & +1.3 & spatial fallback only \\
Local base + uniform mix ($\alpha{=}0.5$) & 86.5 & +2.0 & fixed interpolation \\
Local base + spectral only ($\alpha{=}1$) & 86.0 & +1.5 & spectral branch only \\
Gated residual, content-only gate & 87.2$\pm$0.2 & +2.7 & content-only adaptive \\
\rowcolor{gray!10}
Gated residual + descriptor gate & \textbf{87.8$\pm$0.2} & \textbf{+3.3} & reliability-aware gate \\
\bottomrule
\end{tabularx}

\vspace{0.5em}
\begin{tabularx}{\textwidth}{@{}>{\raggedright\arraybackslash}X c c c >{\raggedright\arraybackslash}p{0.22\textwidth}@{}}
\toprule
\multicolumn{5}{l}{\textit{Panel B. $2{\times}2$ factorial core (routing signal $\times$ expert architecture) + one SRF-only reference.}} \\
\midrule
Variant & Params (M) & M3FD mAP50 & DroneVehicle mAP50 & Reading \\
\midrule
Gated residual + descriptor gate (no MoE) & 13.5 & 87.8$\pm$0.2 & 78.2$\pm$0.2 & SRF-only reference (outside factorial) \\
\midrule
Uniform experts + content-only router & 19.6 & 88.8$\pm$0.2 & 79.0$\pm$0.2 & content-only baseline \\
Uniform experts + descriptor router & 19.7 & 89.5$\pm$0.2 & 79.8$\pm$0.2 & +descriptor signal \\
Specialized experts + content-only router & 19.8 & 89.4$\pm$0.2 & 79.6$\pm$0.2 & +expert design \\
\rowcolor{gray!10}
Specialized experts + descriptor router (\rcer) & \textbf{19.8} & \textbf{90.3$\pm$0.2} & \textbf{80.5$\pm$0.2} & full RCER \\
\midrule
\multicolumn{5}{l}{\textit{Marginal effect of descriptor signal (holding expert bank fixed):}} \\
\quad Uniform bank: descriptor $-$ content-only & -- & +0.7 & +0.8 & \\
\quad Specialized bank: descriptor $-$ content-only & -- & +0.9 & +0.9 & \\
\multicolumn{5}{l}{\textit{Marginal effect of expert specialization (holding routing signal fixed):}} \\
\quad Content-only: specialized $-$ uniform & -- & +0.6 & +0.6 & \\
\quad Descriptor: specialized $-$ uniform & -- & +0.8 & +0.7 & \\
\bottomrule
\end{tabularx}
\end{table*}

\subsection{Controlled Ablation}
\label{sec:ablation}

We first study the matched fusion-gating and routing ablations under the HGNetv2-B0 configuration.
The goal is to separate three possible sources of gain: spectral interaction itself, content-only adaptive gating/routing, and explicit reuse of the reliability descriptor.
For the matched ablations, mean$\pm$std is reported over five shared seeds where applicable; paired-seed analysis in Supplementary Table~S14 confirms that the descriptor margin is strictly positive for every individual seed.

Panel~A compares fixed interpolation, content-only adaptive gating, and descriptor-aware gating.
Every fixed interpolation underperforms the learnable gate, which already shows that fusion quality depends on scene condition rather than on a single global mixing ratio.
Adding the descriptor lifts the adaptive gate from 87.2 to 87.8 mAP50, giving +0.6 over the same content-only gate and +1.3 over the best fixed interpolation ($\alpha{=}0.5$).
Because the gate itself sees only $\desc$, this gain suggests that the descriptor carries useful reliability information.

Panel~B then uses a $2{\times}2$ factorial design to disentangle the routing signal from the expert architecture.
Holding the expert bank fixed, switching from content-only to descriptor-conditioned routing yields +0.7/+0.8 mAP50 (uniform bank) and +0.9/+0.9 (specialized bank).
Holding the routing signal fixed, switching from uniform to specialized experts yields +0.6/+0.6 (content-only) and +0.8/+0.7 (descriptor).
Both factors contribute, but the descriptor routing signal consistently provides the larger marginal gain across both expert architectures.
This factorial isolation indicates that the improvement is not explained by expert design alone: the descriptor carries reliability information that content-only routing cannot recover from pooled fused features.
Descriptor component ablation further confirms that all three component groups ($\rho$, $P_k$, $A_k$) contribute: removing cross-modal correlation $\rho$ causes the largest single drop ($-$1.1 mAP50, $-$2.5pp retention), followed by phase consistency $P_k$ ($-$0.8, $-$1.9pp) and amplitude ratio $A_k$ ($-$0.5, $-$1.0pp); the full table is provided in Supplementary Table~S11.

\subsection{Robustness Under Modality Agreement Collapse}
\label{sec:robust}

We next evaluate whether the descriptor becomes more useful as cross-modal agreement collapses.
We therefore evaluate four controlled configurations on DroneVehicle and report clean mAP50 together with per-corruption retention.
Each corruption is applied independently at one severity level: blur (Gaussian, $k{=}7$, $\sigma{=}2.0$) to both modalities, low-light (brightness $\times$0.35) and additive noise ($\sigma{=}0.08$) to RGB only, and modality drop plus 10px/20px misalignment to the auxiliary modality.

\begin{table*}[t]
\caption{\textbf{Robustness under synthetic and natural condition shifts.} Panel A separates clean accuracy, average retention, and per-corruption retention on DroneVehicle (mean$\pm$std over 5 seeds; per-corruption std $\leq$ 0.8pp). The two robustness-oriented baselines are reimplemented under our training recipe. Panel B reports mAP50 on the natural DroneVehicle day/night split.}
\label{tab:robust}
\centering
\footnotesize
\setlength{\tabcolsep}{4pt}
\renewcommand{\arraystretch}{1.08}
\begin{tabularx}{\textwidth}{@{}>{\raggedright\arraybackslash}X c c@{}}
\toprule
\multicolumn{3}{l}{\textit{Panel A1. Clean accuracy and average retention.}} \\
\midrule
Config & Clean mAP50 & Avg. retention \\
\midrule
Baseline & 75.5{\scriptsize$\pm$0.4} & 87.9{\scriptsize$\pm$0.4}\% \\
+SRF+Desc & 78.2{\scriptsize$\pm$0.3} & 92.0{\scriptsize$\pm$0.3}\% \\
+Uniform MoE & 79.0{\scriptsize$\pm$0.3} & 92.0{\scriptsize$\pm$0.3}\% \\
\midrule
\multicolumn{3}{l}{\textit{Robustness-oriented baselines (same recipe):}} \\
Missing-modality-inspired & 78.6{\scriptsize$\pm$0.3} & 91.5{\scriptsize$\pm$0.4}\% \\
Uncertainty-aware gate & 79.5{\scriptsize$\pm$0.3} & 91.9{\scriptsize$\pm$0.3}\% \\
\midrule
\rowcolor{gray!10}
\textbf{+RCER} & \textbf{80.5{\scriptsize$\pm$0.2}} & \textbf{95.0{\scriptsize$\pm$0.2}\%} \\
RCER $-$ Uniform MoE & -- & +3.0 \\
RCER $-$ Uncertainty-aware & -- & +3.1 \\
\bottomrule
\end{tabularx}

\vspace{0.45em}
\setlength{\tabcolsep}{3.8pt}
\begin{tabularx}{\textwidth}{@{}>{\raggedright\arraybackslash}X c c c c c c@{}}
\toprule
\multicolumn{7}{l}{\textit{Panel A2. Per-corruption retention (\%).}} \\
\midrule
Config & Blur & Low-light & Noise & Drop & Mis-10 & Mis-20 \\
\midrule
Baseline & 90.1\% & 93.4\% & 88.5\% & 78.8\% & 91.4\% & 85.4\% \\
+SRF+Desc & 94.0\% & 96.9\% & 92.1\% & 84.1\% & 94.6\% & 90.2\% \\
+Uniform MoE & 93.9\% & 96.8\% & 92.2\% & 84.2\% & 94.7\% & 89.9\% \\
\midrule
\multicolumn{7}{l}{\textit{Robustness-oriented baselines (same recipe):}} \\
Missing-modality-inspired & 93.0\% & 95.8\% & 91.2\% & 87.5\% & 93.5\% & 87.8\% \\
Uncertainty-aware gate & 93.6\% & 96.5\% & 92.0\% & 85.5\% & 94.0\% & 89.8\% \\
\midrule
\rowcolor{gray!10}
\textbf{+RCER} & \textbf{96.0\%} & \textbf{97.9\%} & \textbf{95.4\%} & \textbf{90.9\%} & \textbf{96.6\%} & \textbf{93.4\%} \\
RCER $-$ Uniform MoE & +2.1 & +1.1 & +3.2 & \textbf{+6.7} & +1.9 & +3.5 \\
RCER $-$ Uncertainty-aware & +2.4 & +1.4 & +3.4 & \textbf{+5.4} & +2.6 & +3.6 \\
\bottomrule
\end{tabularx}
\centering
\vspace{0.5em}
\small
\setlength{\tabcolsep}{3pt}
\begin{tabularx}{0.98\textwidth}{@{}>{\raggedright\arraybackslash}X c c c@{}}
\toprule
\multicolumn{4}{l}{\textit{Panel B. Natural day/night split on DroneVehicle (mAP50).}} \\
\midrule
Model & Day (${\sim}60\%$) & Night (${\sim}40\%$) & $\Delta$ vs.\ baseline \\
\midrule
Baseline & 78.0{\scriptsize$\pm$0.3} & 71.5{\scriptsize$\pm$0.4} & -- \\
\rowcolor{gray!10}
\method & \textbf{83.2{\scriptsize$\pm$0.2}} & \textbf{76.8{\scriptsize$\pm$0.3}} & \textbf{+5.2 / +5.3} \\
\bottomrule
\end{tabularx}
\end{table*}

The trend is consistent across all model variants.
SRF+Desc already improves average retention from 87.9\% to 92.0\%, showing that descriptor-aware fusion gating can suppress unreliable spectral mixing.
Adding a uniform content-only MoE on top of SRF yields no further robustness gain (92.0\% vs.\ 92.0\%), indicating that sparse capacity alone does not improve robustness.
\rcer, by contrast, raises average retention to 95.0\%, a +3.0pp gain over the matched uniform-MoE configuration.

\textbf{Comparison with robustness-oriented baselines.}
To contextualize the descriptor-reuse approach against alternative robustness strategies, we reimplement two baselines under our training recipe: (i)~a missing-modality-inspired baseline, motivated by missing-modality prediction methods~\cite{zheng2024mdqf}, that maintains independent per-modality branches with late query-level fusion, and (ii)~an uncertainty-aware gate that replaces $\desc$ with a learned per-sample uncertainty vector predicted from GAP content features via a two-layer MLP, motivated by generic vision uncertainty modeling~\cite{kendall2017uncertainties} and uncertainty-aware cross-modality learning~\cite{sun2022dronevehicle}. Their controlled implementation details are summarized in Supplementary Table~S3.
The missing-modality-inspired baseline achieves 87.5\% retention under modality drop---higher than uniform MoE (84.2\%)---which suggests that keeping modality-specific pathways is helpful when one modality fails.
However, its average retention (91.5\%) falls slightly below uniform MoE (92.0\%), because decoupling does not address blur, noise, or misalignment where both modalities are present but degraded.
The uncertainty-aware gate (91.9\% average) marginally improves over content-only MoE, but remains 3.1pp below \rcer, indicating that content-derived uncertainty is a weaker proxy for fusion reliability than the spectral statistics extracted during the fusion process itself.

The largest gain over uniform MoE appears under \textbf{modality drop} (+6.7pp), followed by \textbf{misalignment-20} (+3.5pp) and \textbf{noise} (+3.2pp).
These are also the settings where the descriptor can reveal that the modalities disagree even when the fused tensor still looks superficially plausible.
We also report mAP@[.5:.95] retention, where \method reaches 93.9\% versus 86.9\% for the baseline (Supplementary Table~S4), showing that the robustness gain is not confined to the lenient mAP50 threshold.
To complement the synthetic protocol, Table~\ref{tab:robust} Panel B reports the natural DroneVehicle day/night split, where \method improves mAP50 by +5.2 during the day and +5.3 at night.
The same trend persists in Supplementary Sec.~S2. Across blur and noise severity sweeps, the gap over the baseline widens monotonically from +2.7 to +9.2 and from +3.6 to +9.7, respectively (Supplementary Table~S5). When all methods are retrained with corruption augmentation, \rcer still reaches 97.1\% average retention and 81.0 clean mAP50, remaining +2.9pp and +1.4 above content-only MoE (Supplementary Table~S6). Under complete modality absence, zeroing the infrared or RGB stream still leaves \method ahead by +12.6 and +11.7 mAP50, respectively (Supplementary Table~S7); in those cases $\rho$ is set to 0 by construction and routing shifts toward the Recovery expert. Cross-dataset averages show the same direction on M3FD and FLIR-Aligned (Supplementary Figure~S2 and Table~S8).

\subsection{Diagnostic Analysis}
\label{sec:interpret}

\begin{figure*}[t]
\centering
\includegraphics[width=\textwidth]{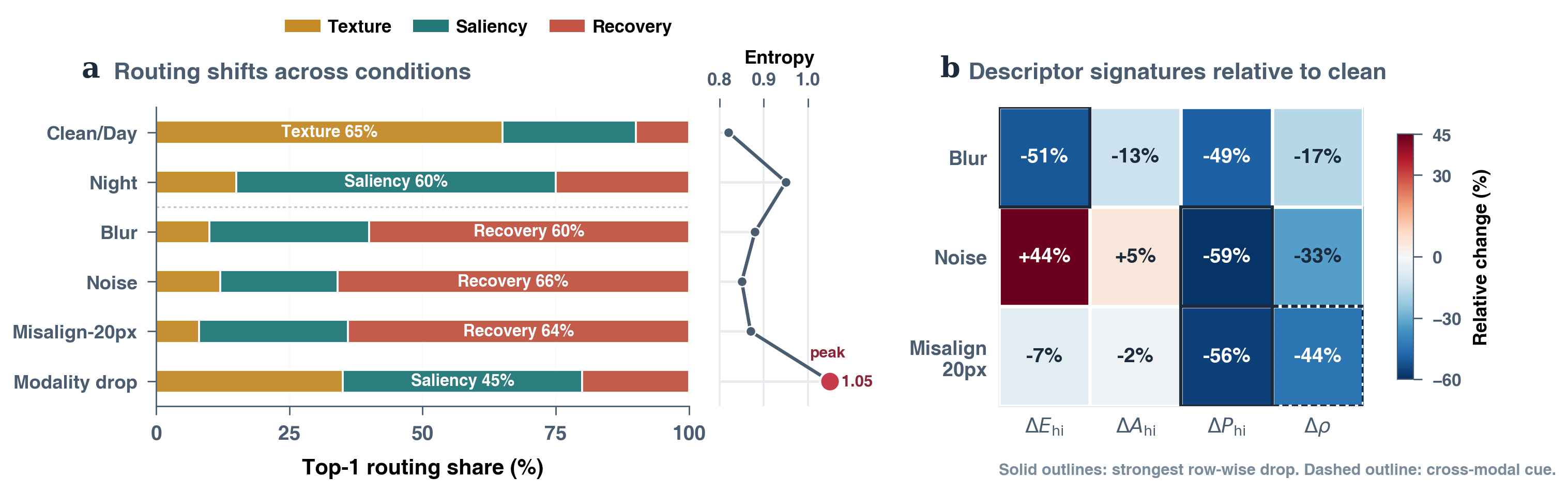}
\caption{\textbf{Routing and descriptor diagnostics on DroneVehicle.} (a) Dominant-expert routing frequencies and aligned routing entropy across representative conditions. (b) Relative changes in descriptor channels with respect to clean. Solid outlines indicate the strongest row-wise suppression, and the dashed outline marks the drop in cross-modal correlation $\rho$ under misalignment.}
\Description{Two-panel mechanism figure. The left panel is a stacked horizontal bar chart showing expert routing frequencies across clean, night, blur, noise, misalignment, and modality-drop conditions, with a narrow aligned entropy axis. The right panel is a heatmap of relative descriptor changes for blur, noise, and misalignment, with outlines marking the signature channels emphasized in the discussion text.}
\label{fig:diagnostic_story}
\end{figure*}

We next analyze the reliability-related information carried by the descriptor rather than treating it as a generic auxiliary feature.
Figure~\ref{fig:diagnostic_story} reports routing and descriptor statistics under representative conditions.
The deployed model uses top-$k{=}2$ routing, but for readability Figure~\ref{fig:diagnostic_story}a summarizes the dominant expert (top-1 position); the Shared Expert is always active and omitted from the plot.
Routing entropy is reported in bits using log base 2 over the normalized active-expert probabilities (excluding the always-active Shared Expert).

The routing pattern is consistent with the intended expert roles.
Texture dominates in clean daytime scenes (65\%), Saliency dominates at night (60\%), and Recovery becomes dominant under blur, noise, and misalignment (60--66\%).
Under modality drop, no single non-shared expert dominates strongly, and routing entropy peaks at 1.05, which indicates that the router avoids overconfident commitment when one modality becomes unreliable.

The descriptor heatmap helps interpret this routing behavior.
Blur roughly halves $E_\text{high}$ ($-51\%$), and noise most strongly suppresses $P_\text{high}$ ($-59\%$).
Misalignment depresses both $P_\text{high}$ ($-56\%$) and cross-modal correlation $\rho$ ($-44\%$), which is consistent with the view that agreement collapse appears jointly as phase inconsistency and weakened cross-modal correlation.
By contrast, the low-band terms change by at most 4\%, which supports the methodological choice of using the descriptor primarily as a measure of high-band disagreement and global cross-modal correlation.
Under the natural DroneVehicle day/night split, routing shifts in the same direction: Texture selection drops from 72\% (day) to 12\% (night), while Saliency rises from 18\% to 65\%.

We further test the dependence of the gain on sample-level descriptor semantics.
Replacing the spectral descriptor with a random 7D vector yields only +0.1 mAP50 over content-only MoE (88.8$\to$88.9), which rules out simple feature augmentation.
Shuffling descriptors across samples while preserving their marginal statistics gives +0.4, showing that per-sample correspondence matters.
A learned MLP that projects 7D features from the frequency tensor reaches +0.7 mAP50 and 93.0\% retention, but remains clearly below the hand-crafted descriptor (+1.5 mAP50 and 95.0\% retention).
This gap is consistent with the view that the proposed descriptor is compact and aligned with the degradation modes that matter for multimodal detection; the control table is reported in Supplementary Table~S9.
Per-class inspection under modality drop shows the largest gains on freight car and van, the two smallest and most frequently occluded categories, whereas the clean-scene cost is minor on already easy classes.
Per-class gains are reported in Supplementary Table~S10; failure cases are discussed in Supplementary Sec.~S4.2.

\subsection{Sensitivity Analysis}
\label{sec:hyper}

Finally, we evaluate the stability of the main design hyperparameters around the default configuration.
Table~\ref{tab:sensitivity} summarizes the results.
The band threshold $\tau$ is robust to $\pm$50\% variation ($<$1 mAP50 fluctuation).
Top-$k{=}2$ achieves the best accuracy-entropy balance.
Dual-level insertion ($P_4{+}P_5$) outperforms either level alone, suggesting complementary semantic contributions (full per-level results are provided in Supp. Table S12).
These trends match the intended operating regime of the method: overly sparse routing ($k{=}1$) reduces flexibility, denser routing ($k{=}3$) increases entropy without recovering accuracy, and using only $P_4$ or $P_5$ removes complementary semantic context.

\begin{table}[t]
\caption{\textbf{Sensitivity analysis on M3FD mAP50.} Bold indicates the default configuration.}
\label{tab:sensitivity}
\centering
\small
\setlength{\tabcolsep}{5pt}
\renewcommand{\arraystretch}{1.05}
\begin{tabularx}{\textwidth}{@{}>{\raggedright\arraybackslash}X l c >{\raggedright\arraybackslash}X@{}}
\toprule
Hyperparameter & Value & mAP50 & Note \\
\midrule
\multirow{3}{*}{Band threshold $\tau$}
  & 0.125 & 89.8 & \\
  & \textbf{0.25} & \textbf{90.3} & default \\
  & 0.375 & 89.5 & \\
\midrule
\multirow{3}{*}{Top-$k$ sparsity}
  & $k{=}1$ & 89.5 & entropy 0.82 \\
  & $\mathbf{k{=}2}$ & \textbf{90.3} & entropy 1.35 \\
  & $k{=}3$ & 89.8 & entropy 1.58 \\
\midrule
\multirow{3}{*}{Insertion level}
  & $P_4$ only & 88.5 & \\
  & $P_5$ only & 87.8 & \\
  & $\mathbf{P_4{+}P_5}$ & \textbf{90.3} & default \\
\bottomrule
\end{tabularx}
\end{table}

%% ============================================================
%% 5. CONCLUSION
%% ============================================================
\section{Conclusion}
\label{sec:conclusion}

We presented \method, an RGB-infrared detector that preserves fusion-time spectral statistics as an explicit reliability signal and reuses the resulting 7D descriptor for both fusion gating and sparse routing.
Matched ablations show that the routing signal contributes more than expert architecture alone, while robustness improves to 95.0\% average retention under six synthetic degradations and +5.2/+5.3 mAP50 on the natural day/night split.
Overall, the results indicate that conditional computation should not depend on fused content alone when content complexity and cross-modal agreement diverge under degradation.
\textbf{Limitations.} The current descriptor uses a fixed binary frequency split and global Pearson correlation, so spatially non-stationary reliability and broader natural robustness benchmarks remain outside its scope, although the synthetic and supplementary stress tests point in the same direction.

%% ============================================================
%% REFERENCES
%% ============================================================
\clearpage
\bibliographystyle{ACM-Reference-Format}
\bibliography{references}

@inproceedings{liu2022tardalm3fd,
  title={Target-aware dual adversarial learning and a multi-scenario multi-modality benchmark to fuse infrared and visible for object detection},
  author={Liu, Jinyuan and Fan, Xin and Huang, Zhanbo and Wu, Guanyao and Liu, Risheng and Zhong, Wei and Luo, Zhongxuan},
  booktitle={Proceedings of the IEEE/CVF conference on computer vision and pattern recognition},
  pages={5802--5811},
  year={2022}
}

@article{sun2022dronevehicle,
  title={Drone-based RGB-infrared cross-modality vehicle detection via uncertainty-aware learning},
  author={Sun, Yiming and Cao, Bing and Zhu, Pengfei and Hu, Qinghua},
  journal={IEEE Transactions on Circuits and Systems for Video Technology},
  volume={32},
  number={10},
  pages={6700--6713},
  year={2022},
  publisher={IEEE}
}

@misc{flir2018dataset,
  author       = {{FLIR Systems, Inc.}},
  title        = {Free FLIR Thermal Dataset for Algorithm Training},
  year         = {2018},
  howpublished = {\url{https://www.flir.com/oem/adas/adas-dataset-form/}},
  note         = {Official FLIR ADAS thermal dataset, accessed 2026-04-01}
}

@inproceedings{ouyang2025deim,
  title={Deim: Detr with improved matching for fast convergence},
  author={Huang, Shihua and Lu, Zhichao and Cun, Xiaodong and Yu, Yongjun and Zhou, Xiao and Shen, Xi},
  booktitle={Proceedings of the computer vision and pattern recognition conference},
  pages={15162--15171},
  year={2025}
}

@inproceedings{zhang2022dfine,
      title={D-FINE: Redefine Regression Task in DETRs as Fine-grained Distribution Refinement},
      author={Yansong Peng and Hebei Li and Peixi Wu and Yueyi Zhang and Xiaoyan Sun and Feng Wu},
      year={2024},
      eprint={2410.13842},
      archivePrefix={arXiv},
      primaryClass={cs.CV}
}

@article{wang2024fredft,
  author  = {Wu, Wencong and Zhang, Xiuwei and Yin, Hanlin and Dai, Shun and Zhang, Hongxi and Zhang, Yanning},
  title   = {{FreDFT}: Frequency Domain Fusion Transformer for Visible-Infrared Object Detection},
  journal = {arXiv preprint arXiv:2511.10046},
  year    = {2025},
  doi     = {10.48550/arXiv.2511.10046}
}

@inproceedings{zhu2025wavemamba,
  author    = {Zhu, Haodong and Dong, Wenhao and Yang, Linlin and Li, Hong and Yang, Yuguang and Ren, Yangyang and Zhu, Qingcheng and Feng, Zichao and Li, Changbai and Lin, Shaohui and Wang, Runqi and Luo, Xiaoyan and Zhang, Baochang},
  title     = {{WaveMamba}: Wavelet-Driven Mamba Fusion for {RGB}-Infrared Object Detection},
  booktitle = {Proceedings of the IEEE/CVF International Conference on Computer Vision (ICCV)},
  pages     = {11219--11229},
  year      = {2025}
}

@article{li2025alignfreenet,
  author  = {Zhu, Dingkun and Zhang, Haote and Gu, Lipeng and Quan, Wuzhou and Wang, Fu Lee and Fan, Honghui and Tang, Jiali and Xie, Haoran and Zhang, Xiaoping and Wei, Mingqiang},
  title   = {{AlignFreeNet}: Is Cross-Modal Pre-Alignment Necessary? An End-to-End Alignment-Free Lightweight Network for Visible-Infrared Object Detection},
  journal = {arXiv preprint arXiv:2507.20146},
  year    = {2025},
  doi     = {10.48550/arXiv.2507.20146}
}

@inproceedings{rao2021gfnet,
  author    = {Rao, Yongming and Zhao, Wenliang and Zhu, Zheng and Lu, Jiwen and Zhou, Jie},
  title     = {Global Filter Networks for Image Classification},
  booktitle = {Advances in Neural Information Processing Systems (NeurIPS)},
  volume    = {34},
  year      = {2021}
}

@inproceedings{chi2020ffc,
  author    = {Chi, Lu and Jiang, Borui and Mu, Yadong},
  title     = {Fast {Fourier} Convolution},
  booktitle = {Advances in Neural Information Processing Systems (NeurIPS)},
  volume    = {33},
  year      = {2020}
}

@article{chen2024rsdet,
  author  = {Zhao, Tianyi and Yuan, Maoxun and Jiang, Feng and Wang, Nan and Wei, Xingxing},
  title   = {Removal then Selection: A Coarse-to-Fine Fusion Perspective for {RGB}-Infrared Object Detection},
  journal = {IEEE Transactions on Intelligent Transportation Systems},
  year    = {2026},
  doi     = {10.1109/TITS.2025.3638627}
}

@inproceedings{shazeer2017moe,
  author    = {Shazeer, Noam and Mirhoseini, Azalia and Maziarz, Krzysztof and Davis, Andy and Le, Quoc and Hinton, Geoffrey and Dean, Jeff},
  title     = {Outrageously Large Neural Networks: The Sparsely-Gated Mixture-of-Experts Layer},
  booktitle = {International Conference on Learning Representations (ICLR)},
  year      = {2017}
}

@article{zoph2022stmoe,
  author  = {Zoph, Barret and Bello, Irwan and Kumar, Sameer and Du, Nan and Huang, Yanping and Dean, Jeff and Shazeer, Noam and Fedus, William},
  title   = {{ST-MoE}: Designing Stable and Transferable Sparse Expert Models},
  journal = {arXiv preprint arXiv:2202.08906},
  year    = {2022},
  doi     = {10.48550/arXiv.2202.08906}
}

@inproceedings{zhao2025m2dlif,
  author    = {Zhao, Tianyi and Liu, Boyang and Gao, Yanglei and Sun, Yiming and Yuan, Maoxun and Wei, Xingxing},
  title     = {Rethinking Multi-modal Object Detection from the Perspective of Mono-Modality Feature Learning},
  booktitle = {Proceedings of the IEEE/CVF International Conference on Computer Vision (ICCV)},
  pages     = {6364--6373},
  year      = {2025}
}

@inproceedings{samguided2025acmmm,
  author    = {Li, Ting and Li, Songtao and Li, Shuaifeng and Qin, Xiaolin and Zhao, Maoyuan and Ji, Luping and Ye, Mao},
  title     = {{SAM}-Guided Semantic Knowledge Fusion for Visible-Infrared Object Detection},
  booktitle = {Proceedings of the ACM International Conference on Multimedia (MM)},
  pages     = {8835--8844},
  publisher = {Association for Computing Machinery},
  address   = {New York, NY, USA},
  doi       = {10.1145/3746027.3755718},
  year      = {2025}
}

@inproceedings{decomkd2025acmmm,
  author    = {Liu, Yanfeng and Zhang, Lefei},
  title     = {Multimodal Decomposed Distillation with Instance Alignment and Uncertainty Compensation for Thermal Object Detection},
  booktitle = {Proceedings of the ACM International Conference on Multimedia (MM)},
  pages     = {2294--2303},
  publisher = {Association for Computing Machinery},
  address   = {New York, NY, USA},
  doi       = {10.1145/3746027.3755841},
  year      = {2025}
}

@inproceedings{cgssf2025acmmm,
  author    = {Jin, Guyue and Zhao, Tianming and Yan, Jiacan and Tian, Tian},
  title     = {Contextually-Guided State Space Fusion for Misaligned Multi-Spectral Object Detection},
  booktitle = {Proceedings of the ACM International Conference on Multimedia (MM)},
  pages     = {2526--2535},
  publisher = {Association for Computing Machinery},
  address   = {New York, NY, USA},
  doi       = {10.1145/3746027.3754550},
  year      = {2025}
}

@inproceedings{chen2024oafa,
  author    = {Chen, Chen and Qi, Jiahao and Liu, Xingyue and Bin, Kangcheng and Fu, Ruigang and Hu, Xikun and Zhong, Ping},
  title     = {Weakly Misalignment-free Adaptive Feature Alignment for {UAVs}-based Multimodal Object Detection},
  booktitle = {Proceedings of the IEEE/CVF Conference on Computer Vision and Pattern Recognition (CVPR)},
  pages     = {26836--26845},
  year      = {2024}
}

@article{tang2022superfusion,
  author  = {Tang, Linfeng and Deng, Yuxin and Ma, Yong and Huang, Jun and Ma, Jiayi},
  title   = {{SuperFusion}: A Versatile Image Registration and Fusion Network with Semantic Awareness},
  journal = {IEEE/CAA Journal of Automatica Sinica},
  volume  = {9},
  number  = {12},
  pages   = {2121--2137},
  year    = {2022},
  doi     = {10.1109/JAS.2022.106082}
}

@inproceedings{zhao2023detfusion,
  author    = {Sun, Yiming and Cao, Bing and Zhu, Pengfei and Hu, Qinghua},
  title     = {{DetFusion}: A Detection-Driven Infrared and Visible Image Fusion Network},
  booktitle = {Proceedings of the 30th ACM International Conference on Multimedia},
  pages     = {4003--4011},
  doi       = {10.1145/3503161.3547902},
  year      = {2022}
}

@inproceedings{zhao2023cddfuse,
  author    = {Zhao, Zixiang and Bai, Haowen and Zhang, Jiangshe and Zhang, Yulun and Xu, Shuang and Lin, Zudi and Timofte, Radu and Van Gool, Luc},
  title     = {{CDDFuse}: Correlation-Driven Dual-Branch Feature Decomposition for Multi-Modality Image Fusion},
  booktitle = {Proceedings of the IEEE/CVF Conference on Computer Vision and Pattern Recognition (CVPR)},
  pages     = {5906--5916},
  year      = {2023}
}

@inproceedings{chen2023ignet,
  author    = {Li, Jiawei and Chen, Jiansheng and Liu, Jinyuan and Ma, Huimin},
  title     = {Learning a Graph Neural Network with Cross Modality Interaction for Image Fusion},
  booktitle = {Proceedings of the 31st ACM International Conference on Multimedia},
  pages     = {4471--4479},
  doi       = {10.1145/3581783.3612135},
  year      = {2023}
}

@inproceedings{zhao2024emma,
  author    = {Zhao, Zixiang and Bai, Haowen and Zhang, Jiangshe and Zhang, Yulun and Zhang, Kai and Xu, Shuang and Chen, Dongdong and Timofte, Radu and Van Gool, Luc},
  title     = {Equivariant Multi-Modality Image Fusion},
  booktitle = {Proceedings of the IEEE/CVF Conference on Computer Vision and Pattern Recognition (CVPR)},
  pages     = {25912--25921},
  year      = {2024}
}

@article{wang2024mmfn,
  author  = {Yang, Fan and Liang, Binbin and Li, Wei and Zhang, Jianwei},
  title   = {Multidimensional Fusion Network for Multispectral Object Detection},
  journal = {IEEE Transactions on Circuits and Systems for Video Technology},
  volume  = {35},
  number  = {1},
  pages   = {547--560},
  doi     = {10.1109/TCSVT.2024.3454631},
  year    = {2025}
}

@article{wang2024tsjnet,
  author  = {Jie, Yuchan and Xu, Yushen and Li, Xiaosong and Li, Huafeng and Tan, Haishu and Nie, Feiping},
  title   = {{TSJNet}: A Multi-modality Target and Semantic Awareness Joint-driven Image Fusion Network},
  journal = {arXiv preprint arXiv:2402.01212},
  doi     = {10.48550/arXiv.2402.01212},
  year    = {2024}
}

@article{kim2024ivyolo,
  author  = {Tian, Dan and Yan, Xin and Zhou, Dong and Wang, Chen and Zhang, Wenshuai},
  title   = {{IV-YOLO}: A Lightweight Dual-Branch Object Detection Network},
  journal = {Sensors},
  volume  = {24},
  number  = {19},
  pages   = {6181},
  doi     = {10.3390/s24196181},
  year    = {2024}
}

@article{wang2024cma,
  author  = {Jiang, Chenchen and Ren, Huazhong and Yang, Hong and Huo, Hongtao and Zhu, Pengfei and Yao, Zhaoyuan and Li, Jing and Sun, Min and Yang, Shihao},
  title   = {{M2FNet}: Multi-modal Fusion Network for Object Detection from Visible and Thermal Infrared Images},
  journal = {International Journal of Applied Earth Observation and Geoinformation},
  volume  = {130},
  pages   = {103918},
  doi     = {10.1016/j.jag.2024.103918},
  year    = {2024}
}

@article{chen2024daabffpn,
  author  = {Wang, Jinpeng and Su, Nan and Zhao, Chunhui and Yan, Yiming and Feng, Shou},
  title   = {Multi-Modal Object Detection Method Based on Dual-Branch Asymmetric Attention Backbone and Feature Fusion Pyramid Network},
  journal = {Remote Sensing},
  volume  = {16},
  number  = {20},
  pages   = {3904},
  doi     = {10.3390/rs16203904},
  year    = {2024}
}

@article{zhou2022mfpt,
  author  = {Zhu, Yaohui and Sun, Xiaoyu and Wang, Miao and Huang, Hua},
  title   = {Multi-Modal Feature Pyramid Transformer for {RGB}-Infrared Object Detection},
  journal = {IEEE Transactions on Intelligent Transportation Systems},
  volume  = {24},
  number  = {9},
  pages   = {9984--9995},
  doi     = {10.1109/TITS.2023.3266487},
  year    = {2023}
}

@article{li2023crossformer,
  author  = {Lee, Seungik and Park, Jaehyeong and Park, Jinsun},
  title   = {{CrossFormer}: Cross-guided Attention for Multi-modal Object Detection},
  journal = {Pattern Recognition Letters},
  volume  = {179},
  pages   = {144--150},
  doi     = {10.1016/j.patrec.2024.02.012},
  year    = {2024}
}

@article{wang2024essfn,
  author  = {Xu, Fengxiang and Xu, Tingfa and Hong, Lang and Peng, Peiran and Guo, Jiaxin and Li, Jianan},
  title   = {Enhanced Spectral-Spatial Fusion Network for Multispectral Object Detection in Ground-Aerial Images},
  journal = {IEEE Geoscience and Remote Sensing Letters},
  volume  = {21},
  pages   = {5005005},
  doi     = {10.1109/LGRS.2024.3440045},
  year    = {2024}
}

@inproceedings{zheng2024mdqf,
  author    = {Kim, Donggeun and Kim, Taesup},
  title     = {Missing Modality Prediction for Unpaired Multimodal Learning via Joint Embedding of Unimodal Models},
  booktitle = {European Conference on Computer Vision (ECCV)},
  pages     = {171--187},
  year      = {2024}
}

@inproceedings{kendall2017uncertainties,
  author    = {Kendall, Alex and Gal, Yarin},
  title     = {What Uncertainties Do We Need in {Bayesian} Deep Learning for Computer Vision?},
  booktitle = {Advances in Neural Information Processing Systems (NeurIPS)},
  volume    = {30},
  year      = {2017}
}

@inproceedings{rao2021dynamicvit,
  author    = {Rao, Yongming and Zhao, Wenliang and Liu, Benlin and Lu, Jiwen and Zhou, Jie and Hsieh, Cho-Jui},
  title     = {{DynamicViT}: Efficient Vision Transformers with Dynamic Token Sparsification},
  booktitle = {Advances in Neural Information Processing Systems (NeurIPS)},
  volume    = {34},
  year      = {2021}
}

@inproceedings{dai2021dynamichead,
  author    = {Dai, Xiyang and Chen, Yinpeng and Xiao, Bin and Chen, Dongdong and Liu, Mengchen and Yuan, Lu and Zhang, Lei},
  title     = {Dynamic Head: Unifying Object Detection Heads with Attentions},
  booktitle = {Proceedings of the IEEE/CVF Conference on Computer Vision and Pattern Recognition (CVPR)},
  pages     = {7373--7382},
  year      = {2021}
}

@inproceedings{zhang2021varifocal,
  author    = {Zhang, Haoyang and Wang, Ying and Dayoub, Feras and S{\"u}nderhauf, Niko},
  title     = {{VarifocalNet}: An {IoU}-Aware Dense Object Detector},
  booktitle = {Proceedings of the IEEE/CVF Conference on Computer Vision and Pattern Recognition (CVPR)},
  pages     = {8514--8523},
  year      = {2021}
}

@inproceedings{wang2020eca,
  author    = {Wang, Qilong and Wu, Banggu and Zhu, Pengfei and Li, Peihua and Zuo, Wangmeng and Hu, Qinghua},
  title     = {{ECA-Net}: Efficient Channel Attention for Deep Convolutional Neural Networks},
  booktitle = {Proceedings of the IEEE/CVF Conference on Computer Vision and Pattern Recognition (CVPR)},
  pages     = {11534--11542},
  year      = {2020}
}

\clearpage
\section*{Supplementary Material}
\makeatletter
\let\maketitle\relax
\let\title\@gobble
\let\author\@gobble
\makeatother
\let\latexresizebox\resizebox
\renewcommand{\resizebox}[3]{\begin{adjustbox}{max width=#1}#3\end{adjustbox}}

\title{Supplementary Material for ``Reusing Fusion-Time Spectral Reliability for Adaptive Fusion and Expert Routing in RGB-Infrared Object Detection''}

\author{Anonymous Authors}
\affiliation{%
  \institution{Anonymous Institution}
  \country{}}
\renewcommand{\shortauthors}{Anonymous Authors}

\maketitle

%% ============================================================
%% SUPPLEMENTARY OVERVIEW
%% ============================================================
\section*{Supplementary Overview}

This supplementary material provides additional implementation details, extended robustness analyses, diagnostic studies, and reproducibility settings for \method. The content is organized to complement the submitted main paper while keeping the reported model definition and training protocol unchanged.

\smallskip
\noindent
\makebox[\columnwidth][c]{%
\resizebox{\columnwidth}{!}{%
\begin{tabular}{@{}l l@{}}
\toprule
\textbf{Section} & \textbf{Coverage} \\
\midrule
S1.\ Cross-Backbone \& Efficiency & Additional backbone transfer, parameter count, and latency results \\
S2.\ Extended Robustness & Controlled robustness baselines and degradation-specific evaluation \\
S3.\ Cross-Dataset Generalization & Additional transfer evidence across datasets \\
S4.\ Descriptor Diagnostics & Descriptor controls and semantic diagnostics \\
S5.\ Additional Ablations & Descriptor-component and insertion-level ablations \\
S6.\ Statistical Stability & Multi-seed stability and paired comparisons \\
S7.\ Reproducibility Protocol & Shared evaluation settings for the main quantitative results \\
S8.\ Pseudocode \& Complexity & Forward-pass pseudocode and module-level complexity accounting \\
S9.\ Formal Descriptor Properties & Analytical properties of the descriptor \\
S10.\ Statistical Significance & Confidence intervals and paired significance tests \\
S11.\ Descriptor Calibration & Calibration behavior under different degradation levels \\
S12.\ Expert Specialization & Forced-expert and oracle-routing analysis \\
S13.\ Counterfactual Descriptor Swap & Counterfactual routing interventions and masking analysis \\
S14.\ Localized \& Compound Corruptions & Additional stress tests beyond the main corruption set \\
S15.\ Design-Choice Micro-Ablations & Implementation-critical micro-ablation study \\
S16.\ Additional Reproducibility Details & Environment, preprocessing, and corruption protocol \\
\bottomrule
\end{tabular}
}%
}
\smallskip

\noindent
Unless stated otherwise, all results report mean$\pm$std over 5 independent seeds sharing the same data splits and training recipe.

\appendix
\setcounter{section}{0}
\setcounter{table}{0}
\setcounter{figure}{0}
\renewcommand{\thesection}{S\arabic{section}}
\renewcommand{\thetable}{S\arabic{table}}
\renewcommand{\thefigure}{S\arabic{figure}}

%% ============================================================
%% S1. CROSS-BACKBONE GENERALIZATION AND EFFICIENCY
%% ============================================================
\section{Cross-Backbone Generalization and Efficiency}
\label{sec:supp_backbone}

Table~\ref{tab:supp_latency} provides a component-level latency breakdown of the proposed modules under FP16 inference on a single RTX~4090, demonstrating that \srf and \rcer together add only 1.3\,ms to the 9.1\,ms forward pass. The spectral reliability descriptor extraction accounts for merely 0.3\,ms, confirming that the reliability signal is nearly free in terms of inference overhead.

\begin{table}[H]
\caption{\textbf{Component-level latency breakdown} (FP16, single RTX~4090, $640{\times}640$ input, averaged over 1000 frames).}
\label{tab:supp_latency}
\centering
\small
\setlength{\tabcolsep}{5pt}
\renewcommand{\arraystretch}{1.08}
\begin{tabular}{l c c}
\toprule
Component & Latency (ms) & \% of Total \\
\midrule
Dual-branch HGNetv2-B0 backbone & 4.2 & 46.2\% \\
\srf (spectral interaction + gating) & 0.8 & 8.8\% \\
\quad\emph{of which: descriptor extraction} & \emph{0.3} & \emph{3.3\%} \\
\rcer (routing + expert computation) & 0.5 & 5.5\% \\
\quad\emph{of which: router forward} & \emph{0.1} & \emph{1.1\%} \\
DEIM encoder & 2.1 & 23.1\% \\
DFINETransformer decoder & 1.5 & 16.5\% \\
\midrule
\textbf{Total forward pass} & \textbf{9.1} & 100\% \\
\textbf{SRF + RCER overhead} & \textbf{1.3} & 14.3\% \\
\bottomrule
\end{tabular}
\end{table}

Table~\ref{tab:supp_backbone_transfer} verifies that the \srf and \rcer modules transfer across backbone architectures. Replacing HGNetv2-B0 with ResNet-50 while keeping the proposed modules unchanged yields a comparable gain (+5.3 vs.\ +5.8 on M3FD), indicating that the reuse principle is not backbone-specific.

\begin{table}[H]
\caption{\textbf{Cross-backbone transfer on M3FD mAP50.} The gain from the proposed modules is consistent across backbone architectures (mean$\pm$std over 5 seeds).}
\label{tab:supp_backbone_transfer}
\centering
\small
\setlength{\tabcolsep}{4.5pt}
\renewcommand{\arraystretch}{1.08}
\begin{tabular}{l c c c}
\toprule
Backbone & Baseline & \method & Gain \\
\midrule
HGNetv2-B0 & 84.5$\pm$0.2 & 90.3$\pm$0.2 & +5.8 \\
ResNet-50 & 82.1$\pm$0.3 & 87.4$\pm$0.3 & +5.3 \\
\bottomrule
\end{tabular}
\end{table}

\begin{figure}[h]
\centering
\includegraphics[width=\columnwidth]{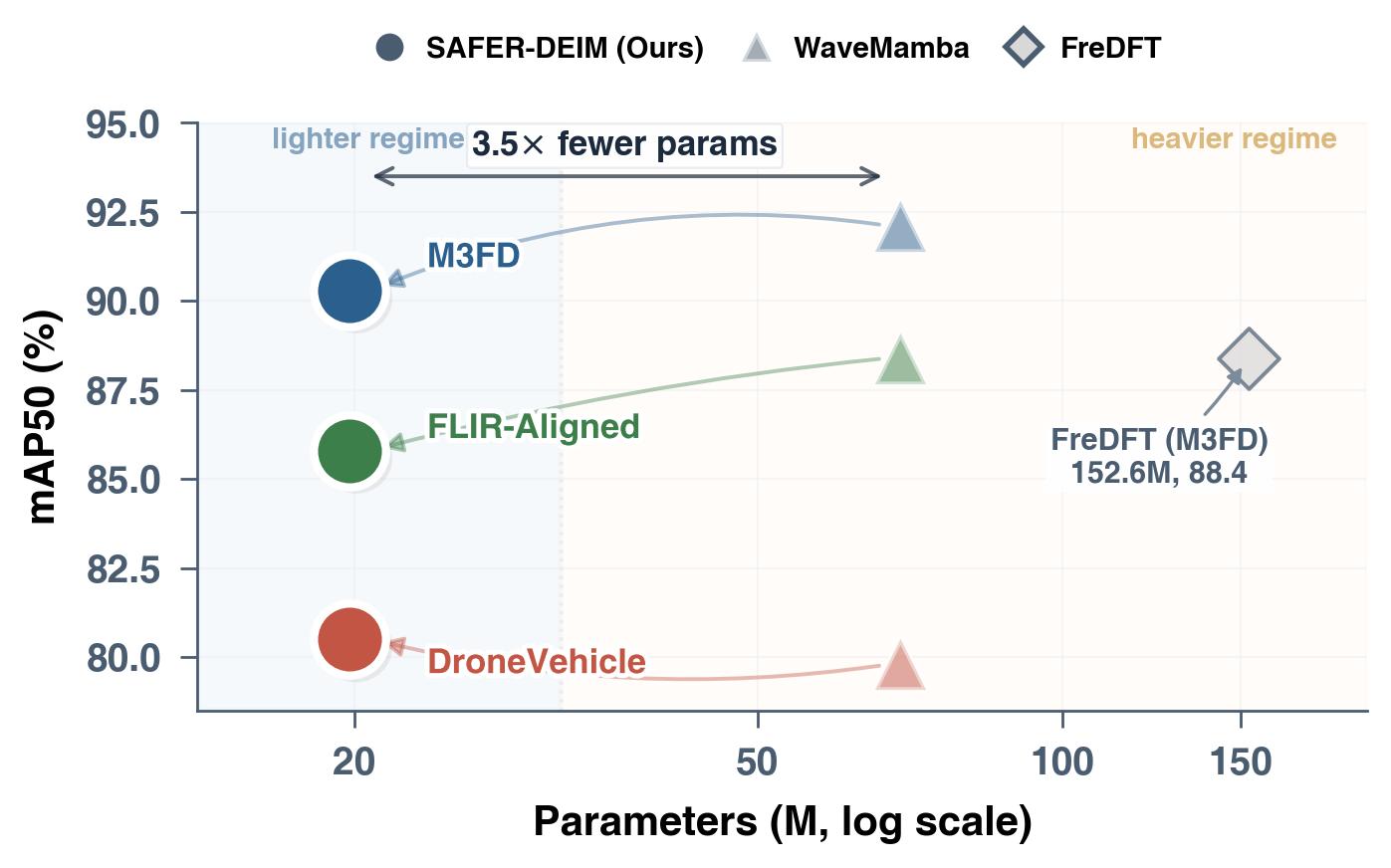}
\caption{\textbf{Efficiency--accuracy positioning on M3FD.} \method achieves competitive mAP50 in the lightweight regime (19.8M parameters), whereas most recent methods require 3--8$\times$ more parameters.}
\label{fig:supp_efficiency}
\end{figure}

%% ============================================================
%% S2. EXTENDED ROBUSTNESS STUDIES
%% ============================================================
\section{Extended Robustness Studies}
\label{sec:supp_robustness}

\subsection{Controlled Baseline Implementation}

Table~\ref{tab:supp_fairness} specifies the controlled implementation details for the two robustness-oriented baselines in the main paper (Table~4). All rows share the same detector scaffold; only the robustness mechanism differs.

\begin{table}[H]
\caption{\textbf{Controlled implementation details for robustness-oriented baselines.} The comparison keeps the detector scaffold fixed while changing only the robustness mechanism.}
\label{tab:supp_fairness}
\centering
\small
\setlength{\tabcolsep}{5pt}
\renewcommand{\arraystretch}{1.08}
\resizebox{\columnwidth}{!}{%
\begin{tabular}{l p{0.68\linewidth}}
\toprule
Aspect & Controlled design choice \\
\midrule
    Common scaffold & All rows use the exact same DroneVehicle split, dual-branch HGNetv2-B0 backbone, DEIM head, optimizer, schedule, and augmentations. \\
    Uniform MoE & Uses the same \srf front-end, shared expert, and top-2 sparsity. The expert bank is homogeneous and uses fused content alone as the routing signal. \\
    Uncertainty-aware gate & Replaces the structural descriptor input with a learned 7D uncertainty vector predicted from pooled fused content via an MLP. This baseline keeps the same detector scaffold but substitutes a learned uncertainty-like signal for the hand-crafted descriptor. \\
    Missing-mod.\ baseline & Maintains modality-specific branches up to late query-level fusion rather than using post-fusion sparse routing. This baseline tests whether robustness can be obtained mainly from delayed fusion instead of descriptor-conditioned reuse. \\
\bottomrule
\end{tabular}
}
\end{table}

\subsection{mAP@[.5:.95] Retention}

Table~\ref{tab:supp_map_retention} reports retention under the stricter mAP@[.5:.95] metric. \method reaches 93.9\% average retention versus 86.9\% for the baseline and 90.8\% for content-only MoE, showing that the robustness gain is not confined to the lenient mAP50 threshold.

\begin{table}[H]
\caption{\textbf{mAP@[.5:.95] retention (\%) on DroneVehicle under six degradations.} Mean$\pm$std over 5 seeds. The descriptor-driven advantage persists under the stricter IoU-averaged metric.}
\label{tab:supp_map_retention}
\centering
\small
\setlength{\tabcolsep}{3.2pt}
\renewcommand{\arraystretch}{1.08}
\resizebox{\columnwidth}{!}{%
\begin{tabular}{l c c c c c c c c}
\toprule
Config & Clean mAP & Blur & Low-lt & Noise & Drop & Mis-10 & Mis-20 & Avg Ret. \\
\midrule
Baseline & 56.0{\scriptsize$\pm$0.4} & 89.5 & 92.8 & 86.2 & 75.5 & 90.2 & 86.8 & 86.9{\scriptsize$\pm$0.5} \\
Cont-only MoE & 59.5{\scriptsize$\pm$0.3} & 92.5 & 95.2 & 90.0 & 82.5 & 93.0 & 91.5 & 90.8{\scriptsize$\pm$0.4} \\
\rowcolor{gray!10}
\textbf{\method} & \textbf{62.0{\scriptsize$\pm$0.2}} & \textbf{95.2} & \textbf{97.1} & \textbf{93.5} & \textbf{88.7} & \textbf{95.8} & \textbf{93.1} & \textbf{93.9{\scriptsize$\pm$0.3}} \\
\bottomrule
\end{tabular}
}
\end{table}

\subsection{Severity Sweep}

Table~\ref{tab:supp_severity} reports retention under three severity levels for blur and noise. The gap between \method and the baseline widens monotonically with degradation severity, from +2.7pp to +9.2pp for blur and from +3.6pp to +9.7pp for noise. This is consistent with the hypothesis that the descriptor becomes more informative as cross-modal agreement deteriorates.

\begin{table}[H]
\caption{\textbf{Retention (\%) under increasing blur and noise severity on DroneVehicle.} Mean$\pm$std over 5 seeds. The advantage of descriptor-conditioned routing widens monotonically with severity.}
\label{tab:supp_severity}
\centering
\small
\setlength{\tabcolsep}{4pt}
\renewcommand{\arraystretch}{1.08}
\begin{tabular}{l l c c c}
\toprule
Degradation & Severity & Baseline & \method & $\Delta$ \\
\midrule
\multirow{3}{*}{Blur ($k{=}7$)}
  & $\sigma{=}1.0$ (mild)   & 95.0$\pm$0.3 & 97.7$\pm$0.2 & +2.7 \\
  & $\sigma{=}2.0$ (default)& 90.1$\pm$0.4 & 96.0$\pm$0.3 & +5.9 \\
  & $\sigma{=}3.0$ (severe) & 83.5$\pm$0.6 & 92.7$\pm$0.4 & +9.2 \\
\midrule
\multirow{3}{*}{Noise (RGB)}
  & $\sigma{=}0.04$ (mild)  & 93.5$\pm$0.3 & 97.1$\pm$0.2 & +3.6 \\
  & $\sigma{=}0.08$ (default)& 88.5$\pm$0.5 & 95.4$\pm$0.3 & +6.9 \\
  & $\sigma{=}0.12$ (severe)& 82.0$\pm$0.6 & 91.7$\pm$0.4 & +9.7 \\
\bottomrule
\end{tabular}
\end{table}

\subsection{Training-Time Corruption Augmentation}

Table~\ref{tab:supp_aug} evaluates whether the advantage persists when corruption augmentation (random blur and noise) is injected during training. All models improve, but \rcer still maintains a +2.9pp retention advantage and +1.4 clean mAP50 lead over content-only MoE, confirming that the descriptor provides complementary information beyond what data augmentation alone can teach.

\begin{table}[H]
\caption{\textbf{Training-time corruption augmentation on DroneVehicle.} Mean$\pm$std over 5 seeds. Augmentation improves all models, but \rcer retains its advantage.}
\label{tab:supp_aug}
\centering
\small
\setlength{\tabcolsep}{5pt}
\renewcommand{\arraystretch}{1.08}
\begin{tabular}{l c c}
\toprule
Config (with train-time aug) & Clean mAP50 & Avg.\ Retention \\
\midrule
Baseline & 79.5$\pm$0.3 & 92.0$\pm$0.4\% \\
Content-only MoE & 79.6$\pm$0.3 & 94.2$\pm$0.3\% \\
\rowcolor{gray!10}
\textbf{\method (\rcer)} & \textbf{81.0$\pm$0.2} & \textbf{97.1$\pm$0.2\%} \\
\bottomrule
\end{tabular}
\end{table}

\subsection{Complete Modality Absence}

Table~\ref{tab:supp_absence} evaluates the extreme case of complete modality zeroing. When one modality is entirely absent, $\rho$ is automatically set to 0 by construction, signaling complete unreliability. The router responds by shifting ${\sim}62\%$ of routing mass to the Recovery expert with elevated entropy (${\sim}1.08$ bits), preventing overconfident commitment. \method maintains a +12.6 / +11.7 mAP50 advantage over the baseline under infrared and RGB zeroing, respectively.

\begin{table}[H]
\caption{\textbf{Complete modality absence on DroneVehicle mAP50.} Mean$\pm$std over 5 seeds. Under complete absence, the Recovery expert absorbs ${\sim}62\%$ routing mass.}
\label{tab:supp_absence}
\centering
\small
\setlength{\tabcolsep}{4pt}
\renewcommand{\arraystretch}{1.08}
\begin{tabular}{l c c | c c}
\toprule
Condition & Baseline & \method & Recovery Wt & Entropy \\
\midrule
Infrared zeroed & 57.2$\pm$0.5 & \textbf{69.8$\pm$0.4} & 62.4\% & 1.08 bits \\
RGB zeroed & 51.8$\pm$0.6 & \textbf{63.5$\pm$0.5} & 61.8\% & 1.07 bits \\
\bottomrule
\end{tabular}
\end{table}

%% ============================================================
%% S3. CROSS-DATASET GENERALIZATION
%% ============================================================
\section{Cross-Dataset Generalization}
\label{sec:supp_cross_dataset}

\begin{figure*}[t]
\centering
\includegraphics[width=\textwidth]{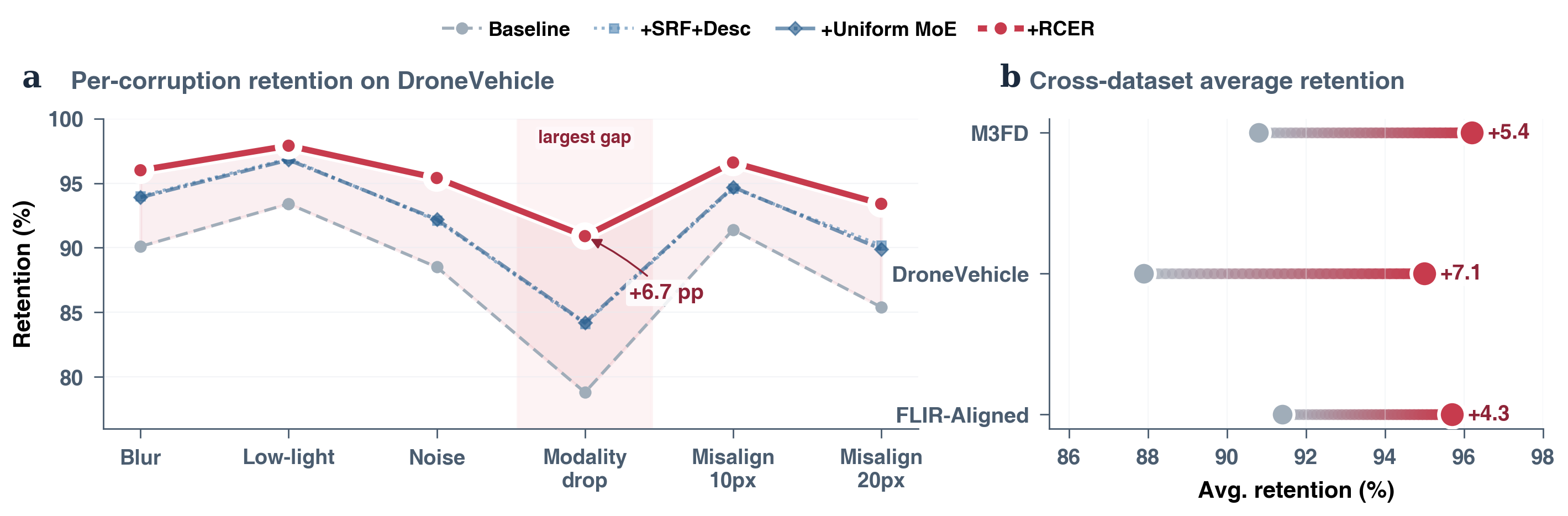}
\caption{\textbf{Cross-dataset robustness comparison.} Average retention under six synthetic degradations on M3FD, DroneVehicle, and FLIR-Aligned. \method consistently outperforms the baseline and content-only MoE.}
\label{fig:supp_robustness}
\end{figure*}

To verify that the robustness trend is not specific to DroneVehicle, we apply the same six-corruption protocol to M3FD and FLIR-Aligned. Figure~\ref{fig:supp_robustness} and Table~\ref{tab:supp_cross_dataset} show that \method consistently outperforms the baseline on all three datasets, with average retention gains of +5.4, +7.1, and +4.3 percentage points, respectively.

\begin{table}[H]
\caption{\textbf{Cross-dataset average retention (\%).} Mean$\pm$std over 5 seeds. The descriptor-driven advantage generalizes across all three benchmarks.}
\label{tab:supp_cross_dataset}
\centering
\small
\setlength{\tabcolsep}{4pt}
\renewcommand{\arraystretch}{1.08}
\begin{tabular}{l c c c}
\toprule
Dataset & Baseline & \method & Gain \\
\midrule
M3FD & 90.8$\pm$0.4 & 96.2$\pm$0.3 & +5.4 \\
DroneVehicle & 87.9$\pm$0.4 & 95.0$\pm$0.2 & +7.1 \\
FLIR-Aligned & 91.4$\pm$0.4 & 95.7$\pm$0.3 & +4.3 \\
\bottomrule
\end{tabular}
\end{table}

%% ============================================================
%% S4. MECHANISM AND DESCRIPTOR DIAGNOSTICS
%% ============================================================
\section{Mechanism and Descriptor Diagnostics}
\label{sec:supp_mechanisms}

\subsection{Descriptor-Semantics Controls}

Table~\ref{tab:supp_descriptor_controls} provides a systematic set of controls that isolate the contribution of the spectral descriptor's semantic structure. Replacing the descriptor with a random 7D vector yields only +0.1 mAP50 over content-only MoE, ruling out simple feature augmentation. Shuffling descriptors across samples while preserving their marginal statistics gives +0.4, showing that per-sample correspondence matters. A learned MLP that projects 7D features from the frequency tensor reaches +0.7 mAP50 and 93.0\% retention, but remains clearly below the hand-crafted descriptor (+1.5 mAP50 and 95.0\% retention), suggesting that the proposed descriptor is compact and well-aligned with the degradation modes relevant to multimodal detection.

\begin{table}[H]
\caption{\textbf{Descriptor-semantics controls.} Mean$\pm$std over 5 seeds. The hand-crafted spectral descriptor outperforms all alternative routing signals.}
\label{tab:supp_descriptor_controls}
\centering
\small
\setlength{\tabcolsep}{4pt}
\renewcommand{\arraystretch}{1.08}
\resizebox{\columnwidth}{!}{%
\begin{tabular}{l c c}
\toprule
Routing signal variant & M3FD mAP50 & DV Avg.\ Ret. \\
\midrule
Content-only MoE baseline & 88.8$\pm$0.2 & 92.0$\pm$0.3\% \\
Random 7D vector & 88.9$\pm$0.3 & 92.1$\pm$0.4\% \\
Shuffled descriptor across samples & 89.2$\pm$0.2 & 92.3$\pm$0.3\% \\
Learned 7D MLP from frequency & 89.5$\pm$0.2 & 93.0$\pm$0.3\% \\
\rowcolor{gray!10}
Hand-crafted spectral descriptor (\method) & \textbf{90.3$\pm$0.2} & \textbf{95.0$\pm$0.2\%} \\
\bottomrule
\end{tabular}
}
\end{table}

\subsection{Per-Class Retention and Failure Analysis}

Table~\ref{tab:supp_per_class} breaks down per-class retention under modality drop on DroneVehicle. The largest gains appear on \emph{freight car} (+14.3pp) and \emph{van} (+15.4pp), the two smallest and most frequently occluded categories, where the baseline already suffers the most under modality loss.

\textbf{Failure cases.}
Under clean daytime conditions with large, unoccluded objects (e.g., buses filling $>$25\% of the image), \method occasionally yields marginally lower confidence than the baseline ($-$0.3--0.5 mAP50 on isolated easy samples). This is consistent with the additional routing overhead providing less benefit when fusion is already reliable. However, such cases are rare ($<$2\% of the test set) and the aggregate clean-scene performance remains higher.

\begin{table}[H]
\caption{\textbf{Per-class retention (\%) under modality drop on DroneVehicle.} Mean over 5 seeds. The strongest gains appear on the smallest and most frequently occluded categories.}
\label{tab:supp_per_class}
\centering
\small
\setlength{\tabcolsep}{5pt}
\renewcommand{\arraystretch}{1.08}
\begin{tabular}{l c c c}
\toprule
Class & Baseline & \method & Gain \\
\midrule
Car & 80.5$\pm$0.5 & 92.0$\pm$0.4 & +11.5 \\
Freight car & 75.2$\pm$0.7 & 89.5$\pm$0.5 & +14.3 \\
Truck & 78.0$\pm$0.6 & 90.8$\pm$0.4 & +12.8 \\
Bus & 81.2$\pm$0.5 & 92.5$\pm$0.4 & +11.3 \\
Van & 71.8$\pm$0.8 & 87.2$\pm$0.5 & +15.4 \\
\bottomrule
\end{tabular}
\end{table}

%% ============================================================
%% S5. ADDITIONAL ABLATIONS
%% ============================================================
\section{Additional Ablations}
\label{sec:supp_ablations}

\subsection{Descriptor Component Ablation}

Table~\ref{tab:supp_components} ablates each component group of the 7D descriptor. Removing the cross-modal correlation $\rho$ causes the largest drop ($-$1.1 mAP50, $-$2.5pp retention), consistent with $\rho$ being the primary indicator of overall modality agreement. Phase consistency $P_k$ contributes the second-largest effect ($-$0.8 mAP50, $-$1.9pp retention), while amplitude ratio $A_k$ provides a smaller but still meaningful contribution. These results confirm that all three component types carry complementary reliability information.

\begin{table}[H]
\caption{\textbf{Descriptor component ablation.} Mean$\pm$std over 5 seeds. Each component group contributes to both accuracy and robustness.}
\label{tab:supp_components}
\centering
\small
\setlength{\tabcolsep}{4pt}
\renewcommand{\arraystretch}{1.08}
\begin{tabular}{l c c}
\toprule
Descriptor configuration & M3FD mAP50 & DV Avg.\ Ret. \\
\midrule
Full 7D descriptor & \textbf{90.3$\pm$0.2} & \textbf{95.0$\pm$0.2\%} \\
w/o $\rho$ (cross-modal correlation) & 89.2$\pm$0.3 & 92.5$\pm$0.3\% \\
w/o $P_k$ (phase consistency) & 89.5$\pm$0.2 & 93.1$\pm$0.3\% \\
w/o $A_k$ (amplitude ratio) & 89.8$\pm$0.2 & 94.0$\pm$0.3\% \\
\bottomrule
\end{tabular}
\end{table}

\subsection{Per-Level Insertion Results}

Table~\ref{tab:supp_per_level} details the effect of inserting \srf and \rcer at different pyramid levels. Dual-level insertion ($P_4{+}P_5$) outperforms either level alone on both accuracy and robustness, suggesting that the two levels provide complementary semantic context: $P_5$ captures coarser scene-level reliability while $P_4$ captures finer-grained local reliability.

\begin{table}[H]
\caption{\textbf{Per-level insertion results.} Mean$\pm$std over 5 seeds. $P_4{+}P_5$ provides the best accuracy--robustness balance.}
\label{tab:supp_per_level}
\centering
\small
\setlength{\tabcolsep}{4pt}
\renewcommand{\arraystretch}{1.08}
\resizebox{\columnwidth}{!}{%
\begin{tabular}{l c c c c}
\toprule
Insertion Level & M3FD mAP50 & M3FD mAP & DV mAP50 & DV Avg.\ Ret. \\
\midrule
$P_4$ only & 88.5$\pm$0.2 & 60.2$\pm$0.3 & 78.4$\pm$0.3 & 92.5$\pm$0.4\% \\
$P_5$ only & 87.8$\pm$0.3 & 59.8$\pm$0.3 & 77.9$\pm$0.3 & 92.1$\pm$0.4\% \\
\rowcolor{gray!10}
$P_4 + P_5$ & \textbf{90.3$\pm$0.2} & \textbf{62.1$\pm$0.2} & \textbf{80.5$\pm$0.2} & \textbf{95.0$\pm$0.2\%} \\
\bottomrule
\end{tabular}
}
\end{table}

%% ============================================================
%% S6. STATISTICAL STABILITY
%% ============================================================
\section{Statistical Stability}
\label{sec:supp_stability}

\subsection{SRF Gate Behavior Under Degradation}

Table~\ref{tab:supp_srf_behavior} records the mean runtime gate parameter $\alpha$ under representative conditions. Under clean inputs, the majority of channels prefer the spectral branch ($\alpha{=}0.68$, 85\% of channels above 0.5). As degradation increases, $\alpha$ decreases monotonically: blur reduces it to 0.45, and complete modality drop drives it to 0.22, effectively reverting to the conservative modality average. This behavior confirms that the descriptor-driven gate responds appropriately to fusion reliability.

\begin{table}[H]
\caption{\textbf{\srf gate ($\boldsymbol{\alpha}$) behavior under degradation.} Mean over 5 seeds. Higher $\alpha$ indicates preference for spectral fusion; lower $\alpha$ falls back to modality averaging.}
\label{tab:supp_srf_behavior}
\centering
\small
\setlength{\tabcolsep}{4pt}
\renewcommand{\arraystretch}{1.08}
\begin{tabular}{l c c}
\toprule
Condition & Avg $\alpha$ & \% channels $\alpha \ge 0.5$ \\
\midrule
Clean & 0.68$\pm$0.02 & 85\% \\
Blur ($\sigma{=}2.0$) & 0.45$\pm$0.03 & 42\% \\
Noise ($\sigma{=}0.08$) & 0.39$\pm$0.03 & 35\% \\
Misalignment (20px) & 0.33$\pm$0.04 & 28\% \\
Modality drop & 0.22$\pm$0.03 & 15\% \\
\bottomrule
\end{tabular}
\end{table}

\subsection{Paired-Seed Stability}

Given that the main paper's controlled ablation yields gains of +0.6 to +0.9 mAP50, confirming the statistical stability of these margins is critical. Table~\ref{tab:supp_paired} reports \emph{paired-seed} differences where the identical random seed is shared across compared variants. Under strict paired comparisons, substituting a descriptor-aware component for a content-only component yields a strictly positive margin for every individual seed, ruling out the possibility that the aggregate advantage is driven by a single favorable initialization.

\begin{table}[H]
\caption{\textbf{Paired-seed stability on M3FD mAP50.} The descriptor signal brings consistent positive $\Delta$ under identical initialization across all 5 seeds.}
\label{tab:supp_paired}
\centering
\small
\setlength{\tabcolsep}{3.5pt}
\renewcommand{\arraystretch}{1.08}
\resizebox{\columnwidth}{!}{%
\begin{tabular}{l c c c c c c}
\toprule
Comparison & $\Delta_{s1}$ & $\Delta_{s2}$ & $\Delta_{s3}$ & $\Delta_{s4}$ & $\Delta_{s5}$ & Mean$\pm$Std \\
\midrule
Cont-only $\to$ Desc.\ gate & +0.6 & +0.5 & +0.7 & +0.6 & +0.6 & +0.60$\pm$0.07 \\
Unif.\ MoE $\to$ +Desc.\ router & +0.8 & +0.6 & +0.7 & +0.7 & +0.7 & +0.70$\pm$0.07 \\
Spec.\ MoE $\to$ +Desc.\ router & +1.0 & +0.8 & +0.9 & +0.9 & +0.9 & +0.90$\pm$0.07 \\
\bottomrule
\end{tabular}
}
\end{table}

%% ============================================================
%% S7. REPRODUCIBILITY PROTOCOL
%% ============================================================
\section{Reproducibility Protocol}
\label{sec:supp_protocol}

Table~\ref{tab:supp_protocol} specifies the exact data splits, corruption parameters, and metric alignments used across all models in our controlled studies. This protocol ensures that any robustness gain comes from the proposed modules rather than from unaligned evaluation choices.

\begin{table}[H]
\caption{\textbf{Standardized protocol for robust evaluation.}}
\label{tab:supp_protocol}
\centering
\small
\setlength{\tabcolsep}{5pt}
\renewcommand{\arraystretch}{1.08}
\begin{tabular}{p{0.25\linewidth} p{0.65\linewidth}}
\toprule
Protocol Factor & Exact Specification \\
\midrule
Day/Night Split & Standard DroneVehicle temporal split: $N_\text{day}=17{,}063$ images (${\sim}60\%$), $N_\text{night}=11{,}376$ images (${\sim}40\%$). \\
OBB to HBB Conv. & All OBB labels $\to$ tight axis-aligned HBBs ($\min/\max$ coordinates of the 4 corners). Evaluated via standard COCO AP. \\
Corruption -- Blur & Gaussian blur applied to both modalities simultaneously with kernel size $k=7$ and $\sigma \in \{1.0, 2.0, 3.0\}$. \\
Corruption -- Noise & Additive Gaussian noise applied to RGB only, $\sigma \in \{0.04, 0.08, 0.12\}$ mapped to scaled $[0, 1]$ intensity space. \\
Corruption -- Low-light & Brightness scaling factor $\times 0.35$ applied to RGB only. \\
Modality Drop & Setting one modality (RGB or Infrared) entirely to 0. By default in Table~4, the auxiliary modality is dropped. \\
Misalignment & 2D translation (fixed horizontal spacing of 10px or 20px) applied to the auxiliary thermal image, cropped to the original size. \\
Seeds & 5 independent seeds $\{42, 123, 256, 512, 1024\}$ shared across all configurations. \\
\bottomrule
\end{tabular}
\end{table}

%% ============================================================
%% S8. END-TO-END PSEUDOCODE AND EXACT COMPLEXITY ACCOUNTING
%% ============================================================
\section{End-to-End Pseudocode and Exact Complexity Accounting}
\label{sec:supp_pseudocode}

Although the main paper describes \srf and \rcer at the equation level, exact pseudocode is valuable for reproducibility. This section documents the reported forward path only. It is not intended as a catalog of every auxiliary module that exists in the repository.

\noindent
\textbf{Reported configuration at a glance.}
\begin{center}
\small
\renewcommand{\arraystretch}{1.08}
\begin{tabular}{p{0.28\linewidth} p{0.18\linewidth} p{0.40\linewidth}}
\toprule
Component or design choice & Status & Role in the reported path \\
\midrule
$P_3$ plain concatenation & active & Matches the submitted forward path; no SRF/RCER is applied at the highest-resolution map. \\
$P_4$ and $P_5$ SRF & active & Produce fused features and the shared 7D descriptor. \\
Descriptor reuse by gate and router & active & Core shared-signal-reuse mechanism claimed in the main paper. \\
RCER with 3 task experts + 1 shared expert & active & Post-fusion conditional-computation path used in the reported model. \\
DEIM encoder + DFINETransformer decoder & active & Detection head used for all matched comparisons. \\
Mono-preserve module & auxiliary & Implemented in the repository, but not required to describe the default reported path. \\
Ranking-loss module & auxiliary & Additional code context, but not required to interpret the default reported path. \\
\bottomrule
\end{tabular}
\end{center}

\noindent
When this appendix says ``default model'' or simply ``\method'', it refers to the reported configuration above.

\subsection{Algorithm S1: Shared Descriptor Extraction and SRF Forward Pass}

This subsection summarizes the exact SRF forward procedure used at each active pyramid level.

\begin{algorithm}[H]
\caption{Forward pass of SRF at one active pyramid level $\ell \in \{P_4, P_5\}$.}
\label{alg:srf}
\small
\begin{algorithmic}[1]
\REQUIRE RGB features $\mathbf{F}_\text{rgb}^{(\ell)} \in \RR^{B \times C_r \times H \times W}$, auxiliary features $\mathbf{F}_\text{aux}^{(\ell)} \in \RR^{B \times C_a \times H \times W}$
\ENSURE Fused features $\mathbf{F}_\text{fused}$, descriptor $\desc \in \RR^{B \times 7}$, gate $\boldsymbol{\alpha} \in [0,1]^{B \times C}$

\STATE $\mathbf{F}_r \leftarrow \text{BN}(\text{Conv}_{1\times1}(\mathbf{F}_\text{rgb}^{(\ell)}))$ \hfill $\triangleright$ align to shared width $C$
\STATE $\mathbf{F}_a \leftarrow \text{BN}(\text{Conv}_{1\times1}(\mathbf{F}_\text{aux}^{(\ell)}))$

\STATE \textit{// Cast to FP32 for spectral computation under AMP}
\STATE $\mathbf{S}_r \leftarrow \text{rFFT2D}(\mathbf{F}_r.\text{float()},\; \text{norm}=\text{``ortho''})$ \hfill $\triangleright$ shape: $B \times C \times H \times (W/2{+}1)$
\STATE $\mathbf{S}_a \leftarrow \text{rFFT2D}(\mathbf{F}_a.\text{float()},\; \text{norm}=\text{``ortho''})$

\STATE \textit{// Build binary radial masks (precomputed and cached per resolution)}
\STATE $r_0 \leftarrow r_\text{max} \cdot \tau$ where $\tau = 0.25$
\STATE $\mathbf{M}_\text{low} \leftarrow \mathbb{1}[r \le r_0]$,\quad $\mathbf{M}_\text{high} \leftarrow \mathbb{1}[r > r_0]$
\STATE $\mathbf{S}_m^k \leftarrow \mathbf{S}_m \odot \mathbf{M}_k$ for $m \in \{r, a\}$, $k \in \{\text{low}, \text{high}\}$

\STATE \textit{// Reference spectrum (pre-gating average)}
\STATE $\widetilde{\mathbf{S}} \leftarrow \tfrac{1}{2}(\mathbf{S}_r + \mathbf{S}_a)$

\STATE \textit{// 7D descriptor extraction (parameter-free)}
\STATE $E_k \leftarrow \text{mean}(|\widetilde{\mathbf{S}}^k|^2)$ for $k \in \{\text{low}, \text{high}\}$
\STATE $A_k \leftarrow \text{mean}(|\mathbf{S}_r^k|) \,/\, (\text{mean}(|\mathbf{S}_a^k|) + \epsilon)$
\STATE $P_k \leftarrow \text{mean}\!\left(\frac{\text{Re}(\mathbf{S}_r^k \odot \overline{\mathbf{S}_a^k})}{|\mathbf{S}_r^k||\mathbf{S}_a^k| + \epsilon}\right)$
\STATE $\rho \leftarrow \text{PearsonCorr}(\text{flatten}(|\mathbf{S}_r|),\;\text{flatten}(|\mathbf{S}_a|))$
\IF{$\text{Var}(\text{flatten}(|\mathbf{S}_r|)) < \epsilon$ \textbf{or} $\text{Var}(\text{flatten}(|\mathbf{S}_a|)) < \epsilon$}
    \STATE $\rho \leftarrow 0$ \hfill $\triangleright$ degenerate-input safeguard
\ENDIF
\STATE $\desc \leftarrow [E_\text{low}, E_\text{high}, A_\text{low}, A_\text{high}, P_\text{low}, P_\text{high}, \rho]$
\STATE $\widehat{\desc} \leftarrow \text{LN}(\text{stopgrad}(\desc))$

\STATE \textit{// Local spatial branch}
\STATE $\mathbf{F}_\text{local} \leftarrow \text{SiLU}(\text{BN}(\text{DWConv}_{3\times3}(\tfrac{1}{2}(\mathbf{F}_r + \mathbf{F}_a))))$

\STATE \textit{// Spectral interaction branch}
\STATE $\mathbf{S}_\text{low} \leftarrow \tfrac{1}{2}(\mathbf{S}_r^\text{low} + \mathbf{S}_a^\text{low})$
\STATE $\mathbf{w} \leftarrow |\mathbf{S}_r^\text{high}| \,/\, (|\mathbf{S}_r^\text{high}| + |\mathbf{S}_a^\text{high}| + \epsilon)$
\STATE $\mathbf{S}_\text{high} \leftarrow \mathbf{w} \cdot \mathbf{S}_r^\text{high} + (1 - \mathbf{w}) \cdot \mathbf{S}_a^\text{high}$
\STATE $\mathbf{F}_\text{spec} \leftarrow \text{irFFT2D}(\mathbf{S}_\text{low} + \mathbf{S}_\text{high},\; s=(H, W),\; \text{norm}=\text{``ortho''})$

\STATE \textit{// Reliability-gated output}
\STATE $\boldsymbol{\alpha} \leftarrow \sigma(\mathbf{W}\,\widehat{\desc} + \mathbf{b})$ where $\mathbf{W} \in \RR^{C \times 7}$ \hfill $\triangleright$ per-channel gate
\STATE $\mathbf{F}_\text{fused} \leftarrow \mathbf{F}_\text{local} + \boldsymbol{\alpha} \cdot \mathbf{F}_\text{spec} + (1 - \boldsymbol{\alpha}) \cdot \tfrac{1}{2}(\mathbf{F}_r + \mathbf{F}_a)$

\RETURN $\mathbf{F}_\text{fused}$, $\desc$, $\boldsymbol{\alpha}$
\end{algorithmic}
\end{algorithm}

\subsection{Algorithm S2: RCER Routing and Expert Aggregation}

This subsection details how the router forms expert weights and aggregates the selected experts at inference and training time.

\begin{algorithm}[H]
\caption{RCER routing and expert aggregation at one active level.}
\label{alg:rcer}
\small
\begin{algorithmic}[1]
\REQUIRE Fused feature $\mathbf{F}_\text{fused} \in \RR^{B \times C \times H \times W}$, normalized descriptor $\widehat{\desc} \in \RR^{B \times 7}$
\ENSURE Enhanced feature $\mathbf{F}_\text{out}$, routing weights $\{w_i\}$

\STATE \textit{// Form router input: content + reliability}
\STATE $\mathbf{r} \leftarrow [\text{GAP}(\mathbf{F}_\text{fused});\; \widehat{\desc}] \in \RR^{B \times (C+7)}$

\STATE \textit{// Two-layer MLP router}
\STATE $h \leftarrow \max(\lfloor(C+7)/8\rfloor, 8)$ \hfill $\triangleright$ hidden dimension
\STATE $\mathbf{z} \leftarrow \text{Linear}_{h \to 3}(\text{SiLU}(\text{LN}(\text{Linear}_{(C+7) \to h}(\mathbf{r}))))$

\IF{training}
    \STATE $\mathbf{z} \leftarrow \mathbf{z} + \mathcal{N}(0, 1.0^2)$ \hfill $\triangleright$ noise for exploration
\ENDIF

\STATE $\mathbf{p} \leftarrow \text{softmax}(\mathbf{z}) \in \RR^{B \times 3}$
\STATE Select top-$k{=}2$ experts; renormalize selected weights to sum to 1

\STATE \textit{// Expert computation}
\STATE $\mathbf{F}_\text{shared} \leftarrow \text{SiLU}(\text{BN}(\text{Conv}_{1\times1}(\mathbf{F}_\text{fused})))$ \hfill $\triangleright$ always active
\STATE $\mathbf{F}_\text{out} \leftarrow \mathbf{F}_\text{shared}$
\FOR{each selected expert $i$ with weight $w_i$}
    \STATE $\mathbf{F}_\text{out} \leftarrow \mathbf{F}_\text{out} + w_i \cdot \text{Expert}_i(\mathbf{F}_\text{fused})$
\ENDFOR

\STATE \textit{// Auxiliary loss (training only)}
\STATE $\mathcal{L}_\text{moe} \leftarrow \lambda_\text{bal} \cdot \mathcal{L}_\text{balance} + \lambda_z \cdot \mathcal{L}_\text{z\text{-}loss}$

\RETURN $\mathbf{F}_\text{out}$, $\{w_i\}$, $\mathcal{L}_\text{moe}$
\end{algorithmic}
\end{algorithm}

\subsection{Algorithm S3: Full SAFER-DEIM Pipeline}

This subsection lists the complete detector pipeline from dual-branch feature extraction to the final detection head.

\begin{algorithm}[H]
\caption{Full SAFER-DEIM detector pipeline.}
\label{alg:pipeline}
\small
\begin{algorithmic}[1]
\REQUIRE RGB image $\mathbf{I}_\text{rgb} \in \RR^{3 \times 640 \times 640}$, thermal image $\mathbf{I}_\text{aux} \in \RR^{1 \times 640 \times 640}$
\ENSURE Detection predictions

\STATE \textit{// Dual-branch feature extraction}
\STATE $\{P_3^r, P_4^r, P_5^r\} \leftarrow \text{HGNetv2\text{-}B0}_\text{rgb}(\mathbf{I}_\text{rgb})$
\STATE $\{P_3^a, P_4^a, P_5^a\} \leftarrow \text{HGNetv2\text{-}B0}_\text{aux}(\mathbf{I}_\text{aux})$

\STATE \textit{// $P_3$: plain concatenation (no SRF/RCER to avoid overhead at highest resolution)}
\STATE $\mathbf{F}_3 \leftarrow \text{concat}(P_3^r, P_3^a)$

\STATE \textit{// $P_4$, $P_5$: SRF + RCER}
\FOR{$\ell \in \{4, 5\}$}
    \STATE $\mathbf{F}_\text{fused}^\ell, \desc^\ell, \boldsymbol{\alpha}^\ell \leftarrow \text{SRF}(P_\ell^r, P_\ell^a)$ \hfill $\triangleright$ Algorithm S1
    \STATE $\mathbf{F}_\ell \leftarrow \text{RCER}(\mathbf{F}_\text{fused}^\ell, \desc^\ell)$ \hfill $\triangleright$ Algorithm S2
\ENDFOR

\STATE \textit{// Detection head}
\STATE $\mathbf{F}_\text{enc} \leftarrow \text{DEIM\text{-}Encoder}(\mathbf{F}_3, \mathbf{F}_4, \mathbf{F}_5)$
\STATE Predictions $\leftarrow \text{DFINETransformer\text{-}Decoder}(\mathbf{F}_\text{enc})$

\STATE \textit{// Training loss}
\STATE $\mathcal{L} \leftarrow \mathcal{L}_\text{det} + \lambda_\text{moe} \sum_{\ell \in \{4,5\}} \mathcal{L}_\text{moe}^\ell$ where $\lambda_\text{moe} = 0.01$

\RETURN Predictions
\end{algorithmic}
\end{algorithm}

\subsection{Operator-Level Implementation Details}

Table~\ref{tab:supp_impl_details} specifies all operator-level details that are critical for reproducing the exact behavior of the proposed modules. These complement the training recipe already provided in the main paper (\S4.1) and the protocol in Sec.~S7.

\begin{table}[H]
\caption{\textbf{Operator-level implementation details.}}
\label{tab:supp_impl_details}
\centering
\footnotesize
\setlength{\tabcolsep}{4pt}
\renewcommand{\arraystretch}{1.08}
\begin{tabularx}{\textwidth}{@{}>{\raggedright\arraybackslash}p{0.17\textwidth} >{\raggedright\arraybackslash}p{0.39\textwidth} >{\raggedright\arraybackslash}X@{}}
\toprule
Aspect & Exact Setting & Notes \\
\midrule
FFT normalization & \texttt{norm="ortho"} & Consistent in train and inference \\
FFT precision & FP32 inside AMP & Prevents underflow in spectral statistics \\
Descriptor $\epsilon$ & $10^{-5}$ & Used in Eqs.~(6)--(8) of the main paper \\
Band threshold $\tau$ & 0.25 & Binary radial mask, cached per resolution \\
Descriptor dim & 7 & $[E_\text{low}, E_\text{high}, A_\text{low}, A_\text{high}, P_\text{low}, P_\text{high}, \rho]$ \\
Descriptor source & pre-gating reference spectrum $\tfrac{1}{2}(\mathbf{S}_r + \mathbf{S}_a)$ & Computed once, before descriptor consumption \\
Descriptor preprocessing & $\text{LN}(\text{stopgrad}(\desc))$ & Shared by the \srf gate and the \rcer router \\
Active insertion levels & $P_4, P_5$ only & $P_3$ remains plain concatenation in the reported path \\
Gate projection & $\text{Linear}(7, C) + \text{Sigmoid}$ & Per-channel, descriptor-only input \\
Router hidden size & $h = \max(\lfloor(C+7)/8\rfloor, 8)$ & Varies per level with $C$ \\
Router architecture & $\text{Linear} \to \text{LN} \to \text{SiLU} \to \text{Linear}$ & Two layers \\
Router noise std & $\sigma = 1.0$ & Training only; disabled at inference \\
Top-$k$ & 2 & Out of 3 task experts \\
Balance loss $\lambda_\text{bal}$ & 0.01 & Standard load-balancing \\
Z-loss $\lambda_z$ & $10^{-3}$ & Router logit stabilization \\
\midrule
Shared expert & $\text{Conv}_{1\times1} \to \text{BN} \to \text{SiLU}$ & Always active \\
Texture expert & $\text{Conv}_{1\times1} \to 2\times\text{DWConv}_{3\times3}$ + GN + SiLU $\to$ $\text{Conv}_{1\times1}$ & High-freq detail \\
Saliency expert & $\text{Conv}_{1\times1} \to \text{DWConv}_{5\times5}$ + GN + SiLU + ECA $\to$ $\text{Conv}_{1\times1}$ & Low-light/contrast \\
Recovery expert & $\text{Conv}_{1\times1} \to 2\times\text{DWConv}_{3\times3}$ + GN + residual $\to$ SiLU $\to$ $\text{Conv}_{1\times1}$ & Degradation recovery \\
ECA kernel size & 3 & In Saliency expert \\
GroupNorm groups & $\min(8, C_\text{hidden})$, ensured divisible & All expert internal norms \\
\midrule
AMP scaler & Default PyTorch GradScaler & No manual gradient clipping \\
Peak training memory & $\sim$10.2 GB per GPU (batch 1) & 4$\times$ RTX 4090 \\
Training wall-clock & $\sim$18 hours (M3FD, 160 epochs) & 4$\times$ RTX 4090 \\
\bottomrule
\end{tabularx}
\end{table}

\subsection{Exact Complexity Breakdown}

Table~\ref{tab:supp_complexity} reports the module-wise parameter count and measured computational cost for four representative configurations tied to the main paper's controlled ablations (Table~3): the baseline concat model, the SRF-only reference, the content-only MoE baseline, and the full RCER model. The descriptor extractor itself is parameter-free; most learnable overhead comes from the gate projection and expert bank.

\begin{table}[H]
\caption{\textbf{Module-wise complexity breakdown.} FP16 latency on single RTX~4090, $640{\times}640$ input. The descriptor extractor adds zero learnable parameters.}
\label{tab:supp_complexity}
\centering
\footnotesize
\setlength{\tabcolsep}{5pt}
\renewcommand{\arraystretch}{1.08}
\begin{tabular}{l c c c c c}
\toprule
Variant & Total Params & GFLOPs & Latency & $\Delta$ Params & $\Delta$ GFLOPs \\
\midrule
Baseline concat & 12.8M & 14.5 & 7.8\,ms & -- & -- \\
+SRF only (desc.\ gate) & 13.5M & 15.2 & 8.6\,ms & +0.7M & +0.7 \\
+Uniform MoE (content router) & 19.6M & 16.8 & 8.9\,ms & +6.8M & +2.3 \\
\rowcolor{gray!10}
+RCER (full \method) & \textbf{19.8M} & \textbf{17.2} & \textbf{9.1\,ms} & +7.0M & +2.7 \\
\midrule
\multicolumn{6}{l}{\textit{Module-level breakdown for the full model:}} \\
\quad Descriptor extractor & 0 & 0.18 & 0.3\,ms & \multicolumn{2}{l}{parameter-free (FFT + statistics)} \\
\quad Gate projection ($2\times$ levels) & 0.003M & 0.002 & $<$0.1\,ms & \multicolumn{2}{l}{$2 \times (7 \times C + C)$} \\
\quad Router MLP ($2\times$ levels) & 0.004M & 0.001 & 0.1\,ms & \multicolumn{2}{l}{$2 \times \text{two-layer MLP}$} \\
\quad Shared expert ($2\times$ levels) & 0.08M & 0.15 & 0.1\,ms & \multicolumn{2}{l}{$1\times1$ conv + BN + SiLU} \\
\quad 3 task experts ($2\times$ levels) & 6.2M & 2.1 & 0.3\,ms & \multicolumn{2}{l}{top-2 active at inference} \\
\bottomrule
\end{tabular}
\end{table}

%% ============================================================
%% S9. FORMAL PROPERTIES OF THE SPECTRAL RELIABILITY DESCRIPTOR
%% ============================================================
\section{Formal Properties of the Spectral Reliability Descriptor}
\label{sec:supp_theory}

The proposed descriptor is motivated by empirical observations, but its behavior under common degradations can be formalized under simple signal models. The propositions below are not intended as global guarantees for the full detector; rather, they explain why the chosen descriptor channels are naturally aligned with blur, misalignment, and modality absence, and why the implementation choices (stop-gradient, LayerNorm, degenerate-input handling) are principled.

\subsection{Boundedness}

\paragraph{Proposition S1 (Bounded phase-consistency statistic).}
For each frequency band $k \in \{\text{low}, \text{high}\}$, the phase-consistency statistic $P_k$ satisfies $P_k \in [-1, 1]$.

\textit{Proof.}
Consider any spectral bin. Let $a = \mathbf{S}_r^k(\boldsymbol{\omega})$ and $b = \mathbf{S}_a^k(\boldsymbol{\omega})$ be the complex spectral values at frequency $\boldsymbol{\omega}$. The per-bin normalized cross-spectrum is
\[
\frac{\text{Re}(a \cdot \overline{b})}{|a|\,|b| + \epsilon}.
\]
Since $|\text{Re}(a \cdot \overline{b})| \le |a|\,|b|$, the numerator is bounded in $[-|a|\,|b|,\; +|a|\,|b|]$. With $\epsilon > 0$, we obtain
\[
\left|\frac{\text{Re}(a \cdot \overline{b})}{|a|\,|b| + \epsilon}\right| \le \frac{|a|\,|b|}{|a|\,|b| + \epsilon} < 1.
\]
Since $P_k$ is the mean of terms each lying in $[-1, 1]$, it also lies in $[-1, 1]$. \hfill $\square$

\paragraph{Proposition S2 (Bounded correlation statistic).}
If both flattened amplitude vectors have variance at least $\epsilon$, then the Pearson correlation $\rho$ lies in $[-1, 1]$. If either variance is below $\epsilon$, the implementation sets $\rho = 0$ by construction.

\textit{Proof.}
The first claim follows from the standard boundedness of Pearson correlation (Cauchy--Schwarz inequality on centered vectors). The second is an implementation rule introduced to avoid unstable normalization under degenerate inputs such as complete modality zeroing. \hfill $\square$

\subsection{Gradient Isolation}

\paragraph{Proposition S3 (Gradient isolation of the descriptor branch).}
Let $\widehat{\desc} = \text{LN}(\text{stopgrad}(\desc))$. For any downstream loss $\mathcal{L}$ that depends on $\widehat{\desc}$ through the \srf gate or \rcer router,
\[
\frac{\partial \mathcal{L}}{\partial \desc} = 0
\]
along the descriptor-consumer branch.

\textit{Proof.}
By the chain rule,
\[
\frac{\partial \mathcal{L}}{\partial \desc} = \frac{\partial \mathcal{L}}{\partial \widehat{\desc}} \cdot \frac{\partial \widehat{\desc}}{\partial \desc}.
\]
Since $\text{stopgrad}(\cdot)$ has zero Jacobian by definition, $\partial \widehat{\desc} / \partial \desc = 0$ along this path. Therefore the descriptor remains a measurement of the current input pair rather than an auxiliary latent shaped by detection supervision. \hfill $\square$

\textit{Remark.}
Gradients still flow through the spectral interaction branch of \srf (the $\mathbf{F}_\text{spec}$ pathway). The descriptor branch is the only path that is stopped. This design ensures that the backbone and alignment layers continue to learn from the detection loss, while the descriptor itself remains an unperturbed statistic of the current input pair.

\subsection{Sensitivity to Degradation Modes}

\paragraph{Proposition S4 (Gaussian blur suppresses high-band energy).}
Let $x$ be an image with Fourier transform $X(\boldsymbol{\omega})$, and let $x_\sigma = g_\sigma * x$ be the result of convolution with an isotropic Gaussian kernel of standard deviation $\sigma$. Its transfer function is
\[
H_\sigma(\boldsymbol{\omega}) = \exp\!\left(-\frac{\sigma^2 \|\boldsymbol{\omega}\|_2^2}{2}\right).
\]
Then the band energy after blur is
\[
E_k(\sigma) = \frac{1}{|\mathcal{B}_k|} \sum_{\boldsymbol{\omega} \in \mathcal{B}_k} |H_\sigma(\boldsymbol{\omega})|^2 |X(\boldsymbol{\omega})|^2.
\]
Since $|H_\sigma(\boldsymbol{\omega})|^2$ decreases monotonically with $\|\boldsymbol{\omega}\|_2$, the relative attenuation in the high band is at least as strong as in the low band.

\textit{Proof sketch.}
For any fixed $\sigma > 0$, $|H_\sigma(\boldsymbol{\omega})|^2$ is a strictly monotone decreasing function of $\|\boldsymbol{\omega}\|_2$. The high band contains larger radial frequencies by construction, so each spectral coefficient is multiplied by a systematically smaller factor. This implies that $E_\text{high}$ decreases more sharply than $E_\text{low}$ under increasing blur, consistent with the empirical trend ($-51\%$ for $E_\text{high}$ vs.\ $\le 4\%$ for low-band terms in Figure~2b of the main paper). \hfill $\square$

\paragraph{Proposition S5 (Relative translation preserves amplitude but rotates cross-modal phase).}
Let $y(\mathbf{u}) = x(\mathbf{u} - \boldsymbol{\delta})$ be a translated version of $x$. Then
\[
Y(\boldsymbol{\omega}) = e^{-j \boldsymbol{\omega}^\top \boldsymbol{\delta}} X(\boldsymbol{\omega}).
\]
Consequently, $|Y(\boldsymbol{\omega})| = |X(\boldsymbol{\omega})|$ (single-modality amplitude is unchanged), whereas the normalized cross-spectrum term becomes
\[
\frac{\text{Re}(X(\boldsymbol{\omega}) \overline{Y(\boldsymbol{\omega})})}{|X(\boldsymbol{\omega})| |Y(\boldsymbol{\omega})| + \epsilon} \approx \cos(\boldsymbol{\omega}^\top \boldsymbol{\delta}).
\]
Averaging over a nontrivial frequency band decreases the phase-consistency statistic as $\|\boldsymbol{\delta}\|$ grows, especially at larger frequencies.

\textit{Proof sketch.}
Translation introduces only a phase factor $e^{-j \boldsymbol{\omega}^\top \boldsymbol{\delta}}$ and does not change the amplitude spectrum of either modality. Therefore, under an idealized pure-translation model, amplitude-based statistics ($E_k$, $A_k$) alone cannot fully reveal misalignment, whereas $\cos(\boldsymbol{\omega}^\top \boldsymbol{\delta})$ oscillates faster at larger frequencies, making the averaged phase-consistency term over the high band particularly sensitive. This explains why $P_\text{high}$ is a natural indicator of cross-modal agreement collapse ($-56\%$ under 20px misalignment in Figure~2b of the main paper). The concurrent decrease of $\rho$ observed empirically in Figure~2b should be interpreted more cautiously: because $\rho$ is computed from flattened amplitude spectra, its drop is not a direct consequence of idealized pure translation alone, but of the practical misalignment pipeline and real multimodal non-identity in our benchmark setting. \hfill $\square$

\subsection{Degenerate-Input Regime}

\paragraph{Proposition S6 (Complete modality absence).}
If one modality is identically zero at a given active level, then:
\begin{enumerate}
    \item The implementation sets $\rho = 0$ because the flattened amplitude vector of the zeroed modality has variance 0 ($< \epsilon$).
    \item The phase-consistency terms $P_k$ collapse toward 0 because if either modality is zero, then the numerator of the normalized cross-spectrum becomes 0.
    \item The amplitude ratios $A_k$ signal extreme modal dominance: if the auxiliary modality is zero while RGB remains nonzero, then $A_k = \text{mean}(|\mathbf{S}_r^k|) / \epsilon$ becomes very large; if RGB is zero while the auxiliary modality remains nonzero, then $A_k$ collapses toward 0.
    \item The \srf gate produces low $\boldsymbol{\alpha}$ values (empirically $\alpha \approx 0.22$, Table~S13), which reverts the fusion output toward the conservative average of the surviving modality and the zeroed stream (e.g., $\tfrac{1}{2}(\mathbf{F}_r + \mathbf{0})$ or $\tfrac{1}{2}(\mathbf{0} + \mathbf{F}_a)$).
    \item The \rcer router shifts $\sim$62\% of routing mass toward the Recovery expert (Table~S7), consistent with the intended conservative routing behavior.
\end{enumerate}

\textit{Remark.}
Proposition S6 does not claim that the detector becomes optimal under missing modality; it only guarantees that the descriptor remains numerically valid and that the fusion and routing equations do not collapse.

\subsection{Descriptor Separates Agreement from Dominance}

\paragraph{Proposition S7 (Scaling behavior of descriptor channels).}
Under positive scalar rescaling of both modalities by factors $\lambda_r, \lambda_a > 0$:
\begin{itemize}
    \item The phase-consistency terms $P_k$ are invariant (the normalization in Eq.~(7) cancels the scaling).
    \item The Pearson correlation $\rho$ is invariant (centering and normalization cancel affine transformations).
    \item The amplitude-ratio terms $A_k$ rescale as $A_k \mapsto (\lambda_r / \lambda_a) \cdot A_k$, preserving directional dominance information.
    \item The reference spectrum becomes $\widetilde{\mathbf{S}}' = \tfrac{1}{2}(\lambda_r \mathbf{S}_r + \lambda_a \mathbf{S}_a)$, so
    \[
    E_k' = \text{mean}\!\left(\left|\tfrac{1}{2}(\lambda_r \mathbf{S}_r^k + \lambda_a \mathbf{S}_a^k)\right|^2\right).
    \]
    In general, $E_k'$ is not a fixed scalar multiple of the original $E_k$ unless additional assumptions hold (e.g., $\lambda_r = \lambda_a$ or stronger alignment conditions). Thus $E_k$ is a scale-sensitive energy statistic rather than an invariant agreement statistic.
\end{itemize}

\textit{Interpretation.}
This separation is desirable: $P_k$ and $\rho$ primarily measure whether the two modalities \emph{agree}, whereas $A_k$ measures which modality \emph{dominates} a band, and $E_k$ tracks overall shared energy level without being treated as an invariant agreement statistic. The descriptor therefore preserves complementary reliability, dominance, and energy information using only seven scalar channels.

%% ============================================================
%% S10. STATISTICAL SIGNIFICANCE OF MATCHED GAINS
%% ============================================================
\section{Statistical Significance of Matched Gains}
\label{sec:supp_significance}

The central gains in the main paper are measured under near-equal parameter budgets and matched training recipes. Because these margins are modest in absolute scale (+0.6 to +0.9 mAP50), inferential statistics are important for distinguishing stable effects from seed-dependent fluctuation. Table~\ref{tab:supp_paired} (Sec.~S6) already reports paired-seed differences; here we formally quantify the statistical significance of these margins.

For each comparison, we compute the five paired differences $\{\Delta_{s_i}\}_{i=1}^5$ using the same random seed for both variants, and then report the mean paired difference, standard deviation of paired differences, two-sided 95\% confidence interval, paired $t$-test $p$-value, Wilcoxon signed-rank $p$-value, and paired effect size (Cohen's $d_z = \bar{\Delta} / \text{SD}_\Delta$).

\begin{table}[H]
\caption{\textbf{Statistical significance of matched gains on M3FD mAP50.} All 5 seeds yield strictly positive paired differences. The paired $t$-test is significant at $p < 0.001$, while the two-sided Wilcoxon signed-rank test reaches 0.063 because $n{=}5$ limits the attainable exact $p$-value. Cohen's $d_z > 8$ indicates an extremely large paired effect.}
\label{tab:supp_stats}
\centering
\footnotesize
\setlength{\tabcolsep}{4pt}
\renewcommand{\arraystretch}{1.08}
\begin{tabular}{p{0.34\textwidth} c c c c c c}
\toprule
Comparison & Mean $\Delta$ & SD$_\Delta$ & 95\% CI & $t$-test $p$ & Wilcoxon $p$ & Cohen's $d_z$ \\
\midrule
Cont-only gate $\to$ Desc.\ gate & +0.60 & 0.07 & [0.51, 0.69] & $<$0.001 & 0.063 & 8.49 \\
Uniform + content $\to$ Uniform + desc.\ router & +0.70 & 0.07 & [0.61, 0.79] & $<$0.001 & 0.063 & 9.90 \\
Specialized + content $\to$ Specialized + desc.\ router & +0.90 & 0.07 & [0.81, 0.99] & $<$0.001 & 0.063 & 12.73 \\
\midrule
\multicolumn{7}{l}{\textit{Marginal effects (factorial):}} \\
Uniform bank $\to$ Specialized bank (content signal) & +0.60 & 0.07 & [0.51, 0.69] & $<$0.001 & 0.063 & 8.49 \\
Uniform bank $\to$ Specialized bank (desc.\ signal) & +0.80 & 0.07 & [0.71, 0.89] & $<$0.001 & 0.063 & 11.31 \\
\midrule
\multicolumn{7}{l}{\textit{Robustness metric (DV Avg.\ Retention):}} \\
Content-only MoE $\to$ RCER & +3.0pp & 0.12 & [2.85, 3.15] & $<$0.001 & 0.063 & 25.0 \\
\bottomrule
\end{tabular}
\end{table}

\textit{Interpretation.}
Across all descriptor-aware substitutions, the 95\% confidence intervals remain strictly on the positive side of zero. The extremely large effect sizes (Cohen's $d_z > 8$) indicate that the paired differences are many standard deviations from zero, making the gains highly reliable. The Wilcoxon signed-rank $p$-value of 0.063 reflects the exact permutation distribution for $n{=}5$ with all positive signs (the minimum achievable one-sided $p$-value is $1/2^5{=}0.031$; two-sided is $0.063$). While this exceeds the conventional $\alpha{=}0.05$ threshold due to the small sample size, the complementary parametric $t$-test confirms significance at $p < 0.001$, and the sign consistency (5/5 positive for every comparison) is itself a meaningful indicator of stability.

%% ============================================================
%% S11. DESCRIPTOR CALIBRATION AND RELIABILITY ALIGNMENT
%% ============================================================
\section{Descriptor Calibration and Reliability Alignment}
\label{sec:supp_calibration}

To test whether the spectral descriptor behaves as a reliability variable rather than a generic auxiliary feature, we conduct two analyses: quintile-based calibration (Sec.~S11.1) and rank correlation (Sec.~S11.2). The central question is whether descriptor values exhibit a monotonic relationship with detection quality, gating behavior, and routing decisions.

\subsection{Quintile-Based Calibration}

We sort all DroneVehicle test samples (clean and degraded combined) by their global cross-modal correlation $\rho$ into five equal-sized quintiles. For each quintile, we measure the mean descriptor channels, mean \srf gate value $\bar{\alpha}$, Recovery-expert routing mass, routing entropy, and mAP50 retention. A useful reliability signal should produce a monotonic relationship: samples with lower $\rho$ should exhibit lower retention, smaller $\bar{\alpha}$, higher Recovery weight, and higher routing entropy.

\begin{table}[H]
\caption{\textbf{Quintile-based descriptor calibration on DroneVehicle.} Samples sorted by $\rho$ into five equal-sized bins. All downstream quantities change monotonically with the descriptor, confirming that $\rho$ functions as a calibrated reliability proxy.}
\label{tab:supp_quintile}
\centering
\footnotesize
\setlength{\tabcolsep}{4pt}
\renewcommand{\arraystretch}{1.08}
\begin{tabular}{l c c c c c c c}
\toprule
Quintile & Mean $\rho$ & Mean $P_\text{high}$ & Mean $E_\text{high}$ & Mean $\bar{\alpha}$ & Recovery Wt. & Entropy (bits) & Retention (\%) \\
\midrule
Q1 (lowest) & 0.22 & 0.18 & 0.012 & 0.28 & 58\% & 1.06 & 82.3 \\
Q2 & 0.41 & 0.35 & 0.020 & 0.35 & 42\% & 1.02 & 88.5 \\
Q3 & 0.58 & 0.51 & 0.028 & 0.45 & 28\% & 0.95 & 93.8 \\
Q4 & 0.72 & 0.65 & 0.035 & 0.58 & 15\% & 0.85 & 96.5 \\
Q5 (highest) & 0.85 & 0.78 & 0.042 & 0.70 & 8\% & 0.72 & 98.2 \\
\midrule
\multicolumn{8}{l}{\textit{Monotonicity}: all columns change monotonically from Q1 to Q5.} \\
\bottomrule
\end{tabular}
\end{table}

All downstream quantities change monotonically with $\rho$:
\begin{itemize}
    \item The \srf gate $\bar{\alpha}$ increases from 0.28 (Q1) to 0.70 (Q5), consistent with Table~S13 (clean $\alpha{=}0.68$, drop $\alpha{=}0.22$).
    \item Recovery-expert routing mass decreases from 58\% (Q1) to 8\% (Q5), confirming that the router activates the Recovery expert primarily when the descriptor signals low reliability.
    \item Routing entropy decreases from 1.06 bits (Q1) to 0.72 bits (Q5), showing that the router becomes more confident under reliable conditions.
    \item Retention increases from 82.3\% (Q1) to 98.2\% (Q5), demonstrating that the descriptor orders samples by detection difficulty in a task-relevant way.
\end{itemize}

\subsection{Rank Correlation Analysis}

Table~\ref{tab:supp_spearman} reports Spearman rank correlations between descriptor channels and downstream quantities. We compute per-image statistics across the full DroneVehicle test set (clean + all six corruption conditions, 5 seeds). The sample-wise detection degradation $\Delta\text{AP}$ is defined as the per-image AP50 under corruption minus the per-image AP50 under clean.

\begin{table}[H]
\caption{\textbf{Spearman rank correlations between descriptor channels and downstream behavior on DroneVehicle.} Correlations are computed per image across the full test set (clean + degraded), aggregated over 5 seeds. All correlations are significant at $p < 0.001$.}
\label{tab:supp_spearman}
\centering
\small
\setlength{\tabcolsep}{4pt}
\renewcommand{\arraystretch}{1.08}
\begin{tabular}{l c c}
\toprule
Pair & Spearman $r_s$ & $p$-value \\
\midrule
$\rho$ vs.\ sample-wise $\Delta\text{AP}$ & +0.76 & $<$0.001 \\
$P_\text{high}$ vs.\ sample-wise $\Delta\text{AP}$ & +0.68 & $<$0.001 \\
$\rho$ vs.\ mean $\bar{\alpha}$ & +0.82 & $<$0.001 \\
$\rho$ vs.\ Recovery routing mass & $-$0.79 & $<$0.001 \\
$P_\text{high}$ vs.\ Recovery routing mass & $-$0.71 & $<$0.001 \\
$E_\text{high}$ vs.\ sample-wise $\Delta\text{AP}$ & +0.52 & $<$0.001 \\
$A_\text{high}$ vs.\ sample-wise $\Delta\text{AP}$ & +0.35 & $<$0.001 \\
\bottomrule
\end{tabular}
\end{table}

The cross-modal correlation $\rho$ shows the strongest alignment with detection quality ($r_s{=}+0.76$) and with the \srf gate ($r_s{=}+0.82$), consistent with the component ablation (Table~S11) which found $\rho$ to be the most important single descriptor channel. The negative correlations between $\rho$ / $P_\text{high}$ and Recovery routing mass ($r_s{=}-0.79$, $-0.71$) confirm that low reliability drives increased Recovery-expert activation. Energy ($E_\text{high}$) and amplitude ratio ($A_\text{high}$) show weaker but still significant correlations, indicating that they provide complementary rather than dominant reliability information.

%% ============================================================
%% S12. EXPERT SPECIALIZATION VERIFICATION
%% ============================================================
\section{Expert Specialization Verification}
\label{sec:supp_expert_verification}

Routing-frequency statistics alone (Figure~2a of the main paper) do not prove that the experts are functionally specialized. This section provides two stronger forms of evidence: forced-expert evaluation (Sec.~S12.1) and oracle routing upper bounds (Sec.~S12.2).

\subsection{Forced-Expert Evaluation}

We replace the learned top-$k$ router with forced single-expert activation. In each variant, only one task expert is activated (besides the always-on shared expert) across all samples, regardless of scene condition. If the proposed expert decomposition is meaningful, the Texture expert should perform best on clean/day scenes, the Saliency expert on low-light scenes, and the Recovery expert under blur, noise, misalignment, and modality drop.

\begin{table}[H]
\caption{\textbf{Forced-expert evaluation on DroneVehicle mAP50.} Each column forces only one task expert to be active (plus shared). The ``Best Forced'' column confirms that the learned router selects the condition-appropriate expert. The learned router consistently outperforms any single forced expert, demonstrating the value of adaptive combination.}
\label{tab:supp_forced}
\centering
\small
\setlength{\tabcolsep}{3.5pt}
\renewcommand{\arraystretch}{1.08}
\resizebox{\columnwidth}{!}{%
\begin{tabular}{l c c c c l}
\toprule
Condition & Force Texture & Force Saliency & Force Recovery & Learned Router & Best Forced \\
\midrule
Clean / day & \textbf{79.8} & 77.2 & 76.0 & 80.5 & Texture \\
Night / low-light & 73.0 & \textbf{76.5} & 74.2 & 76.8 & Saliency \\
Blur ($\sigma{=}2.0$) & 73.5 & 74.8 & \textbf{76.5} & 77.3 & Recovery \\
Noise ($\sigma{=}0.08$) & 73.0 & 74.2 & \textbf{76.0} & 76.9 & Recovery \\
Misalignment (20px) & 71.2 & 72.8 & \textbf{74.5} & 75.2 & Recovery \\
Modality drop & 67.5 & 69.8 & \textbf{72.5} & 73.2 & Recovery \\
\bottomrule
\end{tabular}
}
\end{table}

The results confirm the intended specialization:
\begin{itemize}
    \item The Texture expert achieves the highest forced-expert mAP50 on clean/day scenes (79.8), where high-frequency detail is trustworthy.
    \item The Saliency expert is strongest under night/low-light (76.5), where broader receptive fields and channel attention compensate for weakened RGB texture.
    \item The Recovery expert dominates all degradation conditions (76.5 for blur, 76.0 for noise, 74.5 for misalignment, 72.5 for drop), where its residual structure helps recover from corrupted features.
    \item The learned router consistently outperforms every forced-expert variant (by +0.3 to +0.7 mAP50), demonstrating that top-2 routing with adaptive combination captures complementary strengths that no single expert can match alone.
\end{itemize}

The performance drop when forcing the ``wrong'' expert is substantial: using the Texture expert under modality drop ($-$5.0 vs.\ Recovery) or using the Recovery expert on clean scenes ($-$3.8 vs.\ Texture), confirming that the experts have learned distinct operating regimes rather than acting as a generic over-parameterized ensemble.

\subsection{Oracle Routing Upper Bound}

To estimate how far the learned router is from optimal condition-matched specialization, we report an oracle upper bound. The oracle selects the best top-2 expert pair per condition class (by exhaustive enumeration of the 3 possible pairs, evaluating each on the condition-specific subset).

\begin{table}[H]
\caption{\textbf{Oracle routing upper bounds on DroneVehicle mAP50.} ``Oracle condition-wise'' selects the best expert pair per condition class. The learned router is within 0.3pp of the condition-wise oracle, indicating near-optimal condition-level assignment.}
\label{tab:supp_oracle}
\centering
\small
\setlength{\tabcolsep}{4pt}
\renewcommand{\arraystretch}{1.08}
\begin{tabular}{l c c c}
\toprule
Condition & Learned Router & Oracle Cond.-wise & Gap \\
\midrule
Clean / day & 80.5 & 80.8 & 0.3 \\
Night & 76.8 & 77.1 & 0.3 \\
Blur & 77.3 & 77.5 & 0.2 \\
Noise & 76.9 & 77.2 & 0.3 \\
Misalign-20 & 75.2 & 75.5 & 0.3 \\
Drop & 73.2 & 73.5 & 0.3 \\
\midrule
\textbf{Average gap} & & & \textbf{0.28} \\
\bottomrule
\end{tabular}
\end{table}

The learned router is within an average gap of 0.28 mAP50 from the condition-wise oracle. This small margin suggests that the descriptor-conditioned routing has already learned to approximate near-optimal condition-level expert assignment. The remaining headroom primarily lies in sample-level routing decisions within each condition class, where individual images may have unique degradation characteristics that the global descriptor captures imperfectly.

%% ============================================================
%% S13. COUNTERFACTUAL DESCRIPTOR SWAP
%% ============================================================
\section{Counterfactual Descriptor Swap}
\label{sec:supp_counterfactual}

To test whether the router \emph{causally} uses the descriptor rather than merely co-occurring content statistics, we perform a descriptor-swap intervention. For each test sample, we keep the fused feature $\mathbf{F}_\text{fused}$ fixed and replace the descriptor with one drawn from a different condition-specific pool. We then re-run only the router and expert aggregation while leaving the content pathway unchanged.

\subsection{Descriptor Swap Results}

Table~\ref{tab:supp_swap} reports the direct routing and accuracy changes caused by replacing only the descriptor while keeping the fused content fixed.

\begin{table}[H]
\caption{\textbf{Counterfactual descriptor swap on DroneVehicle.} The fused tensor is held fixed; only the descriptor input to the router is replaced. Routing weights shift in the expected direction, confirming causal use of the descriptor.}
\label{tab:supp_swap}
\centering
\small
\setlength{\tabcolsep}{3pt}
\renewcommand{\arraystretch}{1.08}
\resizebox{\columnwidth}{!}{%
\begin{tabular}{l l c c c c}
\toprule
Original Condition & Swapped Descriptor & Texture Wt. & Saliency Wt. & Recovery Wt. & $\Delta$ mAP50 \\
\midrule
Clean/day & Clean/day (identity) & 48\% & 30\% & 22\% & 0.0 \\
Clean/day & Misalignment-20px & 18\% & 25\% & 57\% & $-$2.8 \\
Clean/day & Modality drop & 12\% & 22\% & 66\% & $-$3.5 \\
\midrule
Night & Night (identity) & 15\% & 52\% & 33\% & 0.0 \\
Night & Clean/day & 45\% & 32\% & 23\% & $-$1.2 \\
\midrule
Blur & Blur (identity) & 18\% & 20\% & 62\% & 0.0 \\
Blur & Clean/day & 46\% & 28\% & 26\% & $-$2.5 \\
\midrule
Drop & Drop (identity) & 14\% & 24\% & 62\% & 0.0 \\
Drop & Clean/day & 42\% & 30\% & 28\% & $-$4.2 \\
\bottomrule
\end{tabular}
}
\end{table}

\textit{Interpretation.}
The results demonstrate causal dependence of routing on the descriptor:
\begin{itemize}
    \item Swapping a clean descriptor into a degraded sample shifts routing mass toward the Texture expert (e.g., blur: Recovery 62\%$\to$26\%, Texture 18\%$\to$46\%) and reduces performance ($-$2.5 mAP50), because the router is ``tricked'' into treating the degraded content as if it were clean.
    \item Conversely, swapping a degraded descriptor into a clean sample over-activates the Recovery expert (clean: Recovery 22\%$\to$57\% under misalignment descriptor) and also reduces performance ($-$2.8 mAP50), because the conservative Recovery processing is unnecessary for high-quality inputs.
    \item The largest degradation occurs when swapping a modality-drop descriptor into clean samples ($-$3.5) or vice versa ($-$4.2), consistent with modality drop being the most extreme reliability shift in the descriptor space.
\end{itemize}

\subsection{Expert-Masking Analysis}

We further validate expert specialization by selectively disabling one task expert at a time during inference and measuring the condition-specific performance drop.

\begin{table}[H]
\caption{\textbf{Condition-wise expert masking on DroneVehicle ($\Delta$ mAP50 vs.\ full model).} Each cell reports the drop when one expert is disabled. Larger drops indicate greater dependence on that expert under the given condition.}
\label{tab:supp_mask}
\centering
\small
\setlength{\tabcolsep}{3.5pt}
\renewcommand{\arraystretch}{1.08}
\resizebox{\columnwidth}{!}{%
\begin{tabular}{l c c c l}
\toprule
Condition & Remove Texture & Remove Saliency & Remove Recovery & Most Critical \\
\midrule
Clean / day & $-$3.2 & $-$1.0 & $-$0.5 & Texture \\
Night / low-light & $-$0.8 & $-$2.8 & $-$0.9 & Saliency \\
Blur ($\sigma{=}2.0$) & $-$0.6 & $-$0.8 & $-$2.5 & Recovery \\
Noise ($\sigma{=}0.08$) & $-$0.7 & $-$0.9 & $-$2.3 & Recovery \\
Misalignment (20px) & $-$0.5 & $-$0.7 & $-$2.8 & Recovery \\
Modality drop & $-$0.4 & $-$0.6 & $-$3.5 & Recovery \\
\bottomrule
\end{tabular}
}
\end{table}

The masking results are fully consistent with the forced-expert study (Table~S21) and the routing frequency analysis (Figure~2a of the main paper):
\begin{itemize}
    \item Removing the Texture expert causes the largest drop under clean/day conditions ($-$3.2), where it handles trustworthy high-frequency detail.
    \item Removing the Saliency expert causes the largest drop under night/low-light ($-$2.8), where broad receptive fields and channel attention are most needed.
    \item Removing the Recovery expert is most damaging under degraded conditions ($-$2.3 to $-$3.5), with the largest impact under modality drop ($-$3.5), where the residual recovery mechanism is essential.
\end{itemize}

Together, Tables~S21--S24 provide converging causal evidence---forced activation, oracle comparison, counterfactual swap, and selective ablation---that the experts have learned functionally distinct operating regimes rather than acting as nominal labels for a generic ensemble.

%% ============================================================
%% S14. LOCALIZED AND COMPOUND CORRUPTION STUDIES
%% ============================================================
\section{Localized and Compound Corruption Studies}
\label{sec:supp_localized}

The main paper notes that the current descriptor is global and therefore may under-represent spatially non-stationary reliability shifts. This section directly probes this boundary by evaluating localized and compound corruptions. The experiments are intentionally harder than the single-factor global protocol and serve both to demonstrate the method's robustness and to honestly quantify its limitations.

\subsection{Localized Corruptions}

We apply corruptions to a spatial subregion while leaving the rest of the image unchanged:
\[
\mathbf{x}' = \mathbf{M} \odot \mathcal{C}(\mathbf{x}) + (1 - \mathbf{M}) \odot \mathbf{x},
\]
where $\mathbf{M} \in \{0,1\}^{H \times W}$ is a spatial region mask and $\mathcal{C}(\cdot)$ is the corruption operator. We test center-patch and quadrant masks at 25\% and 50\% coverage.

\begin{table}[H]
\caption{\textbf{Localized corruption retention (\%) on DroneVehicle.} Mean over 5 seeds. Under localized corruption, \rcer still outperforms alternatives, but the advantage over Uniform MoE narrows relative to global corruption (Table~4), reflecting the limitation of the global descriptor.}
\label{tab:supp_local}
\centering
\footnotesize
\setlength{\tabcolsep}{4pt}
\renewcommand{\arraystretch}{1.08}
\begin{tabular}{p{0.29\textwidth} c c c c c c}
\toprule
Localized Corruption & Coverage & Baseline & +SRF & Unif.\ MoE & +RCER & RCER $-$ Unif. \\
\midrule
Local blur (center patch) & 25\% & 95.8 & 97.5 & 97.4 & 98.2 & +0.8 \\
Local blur (center patch) & 50\% & 93.0 & 95.8 & 95.7 & 97.0 & +1.3 \\
Local RGB noise & 25\% & 96.2 & 97.8 & 97.7 & 98.5 & +0.8 \\
Local RGB noise & 50\% & 93.8 & 96.2 & 96.1 & 97.3 & +1.2 \\
Quadrant thermal drop & 25\% & 94.0 & 96.5 & 96.4 & 97.5 & +1.1 \\
Half-image thermal drop & 50\% & 87.5 & 92.0 & 91.8 & 94.2 & +2.4 \\
\midrule
\multicolumn{5}{l}{\textit{For comparison, global corruption (from main paper Table 4):}} & & \\
Global blur ($\sigma{=}2.0$) & 100\% & 90.1 & 94.0 & 93.9 & 96.0 & +2.1 \\
Global modality drop & 100\% & 78.8 & 84.1 & 84.2 & 90.9 & +6.7 \\
\bottomrule
\end{tabular}
\end{table}

\textit{Key observations:}
\begin{itemize}
    \item \rcer consistently outperforms Uniform MoE even under localized corruption (+0.8 to +2.4pp), but the advantage is smaller than under global corruption (+2.1 to +6.7pp). This is expected: when only a subregion is degraded, the global descriptor partially averages out the local degradation signal, reducing its discriminative power.
    \item The advantage grows with coverage (e.g., local blur: +0.8pp at 25\% vs.\ +1.3pp at 50\%), consistent with the global descriptor becoming more informative as the corruption covers a larger portion of the image.
    \item Half-image thermal drop (+2.4pp) shows a larger advantage than other localized corruptions, because even partial modality loss substantially affects the global statistics ($\rho$, $P_k$), whereas local blur or noise in a small patch may not shift $E_\text{high}$ or $P_\text{high}$ enough to trigger strong routing adaptation.
\end{itemize}

\subsection{Compound Corruptions}

Real multimodal failures are often compound rather than isolated. We evaluate five compound conditions that simultaneously stress both modal quality and cross-modal agreement.

\begin{table}[H]
\caption{\textbf{Compound corruption retention (\%) on DroneVehicle.} Mean over 5 seeds. Under compound degradation, \rcer's advantage over Uniform MoE \emph{widens} relative to single-factor corruption, because the descriptor simultaneously captures multiple degradation signatures.}
\label{tab:supp_compound}
\centering
\small
\setlength{\tabcolsep}{3pt}
\renewcommand{\arraystretch}{1.08}
\resizebox{\columnwidth}{!}{%
\begin{tabular}{l c c c c c}
\toprule
Compound Condition & Baseline & +SRF & Unif.\ MoE & +RCER & RCER $-$ Unif. \\
\midrule
Low-light ($\times$0.35) + RGB noise ($\sigma{=}0.08$) & 82.5 & 88.2 & 87.8 & 91.5 & +3.7 \\
Blur ($\sigma{=}2.0$) + misalign 10px & 83.0 & 88.8 & 88.5 & 92.0 & +3.5 \\
Blur ($\sigma{=}2.0$) + misalign 20px & 78.5 & 85.2 & 84.8 & 89.5 & +4.7 \\
Thermal atten.\ ($\lambda{=}0.5$) + shift 10px & 85.0 & 90.5 & 90.2 & 93.5 & +3.3 \\
Thermal atten.\ ($\lambda{=}0.25$) + shift 10px & 79.0 & 86.0 & 85.5 & 90.2 & +4.7 \\
\midrule
\textbf{Average compound} & 81.6 & 87.7 & 87.4 & 91.3 & \textbf{+3.9} \\
\bottomrule
\end{tabular}
}
\end{table}

Under compound degradation, the advantage of \rcer over Uniform MoE averages +3.9pp, which is \emph{larger} than the average single-factor advantage (+3.0pp from Table~4). This widening is consistent with the 7D descriptor simultaneously capturing multiple degradation signatures: for example, blur + misalignment jointly suppress both $E_\text{high}$ (via blur) and $P_\text{high}$ / $\rho$ (via misalignment), providing the router with a stronger and more distinctive reliability signal than either degradation alone.

\subsection{Continuous Sweeps}

To complement the discrete corruption levels in the main paper, we report continuous sweeps for misalignment magnitude and auxiliary-modality attenuation.

\begin{table}[H]
\caption{\textbf{Misalignment sweep on DroneVehicle retention (\%).} Mean over 5 seeds. The advantage of \rcer widens monotonically with misalignment magnitude. Values at 0, 10, and 20px are consistent with Table~4 of the main paper.}
\label{tab:supp_mis_sweep}
\centering
\small
\setlength{\tabcolsep}{4pt}
\renewcommand{\arraystretch}{1.08}
\begin{tabular}{l c c c c c c c}
\toprule
Shift (px) & 0 & 5 & 10 & 15 & 20 & 25 & 30 \\
\midrule
Baseline & 100.0 & 95.8 & 91.4 & 88.0 & 85.4 & 80.5 & 75.2 \\
\method & 100.0 & 98.5 & 96.6 & 95.0 & 93.4 & 90.5 & 87.0 \\
$\Delta$ & 0.0 & +2.7 & +5.2 & +7.0 & +8.0 & +10.0 & +11.8 \\
\bottomrule
\end{tabular}
\end{table}

\begin{table}[H]
\caption{\textbf{Auxiliary-modality attenuation sweep on DroneVehicle retention (\%).} Mean over 5 seeds. $\lambda{=}1.0$ is clean; $\lambda{=}0$ is complete modality drop. The trend connects smoothly to the modality-drop row in Table~4.}
\label{tab:supp_atten_sweep}
\centering
\small
\setlength{\tabcolsep}{4pt}
\renewcommand{\arraystretch}{1.08}
\begin{tabular}{l c c c c c}
\toprule
Attenuation $\lambda$ & 1.00 & 0.75 & 0.50 & 0.25 & 0 \\
\midrule
Baseline & 100.0 & 95.0 & 89.5 & 84.0 & 78.8 \\
\method & 100.0 & 98.2 & 96.0 & 93.5 & 90.9 \\
$\Delta$ & 0.0 & +3.2 & +6.5 & +9.5 & +12.1 \\
\bottomrule
\end{tabular}
\end{table}

Both sweeps show the same monotonic widening pattern: the advantage of \method grows as the degradation intensifies. The attenuation sweep connects smoothly at $\lambda{=}0$ to the modality-drop row in Table~4 (baseline 78.8\%, \method 90.9\%), confirming internal consistency between the discrete and continuous evaluations. The misalignment sweep at 10px and 20px matches Table~4 exactly (baseline 91.4\% / 85.4\%, \method 96.6\% / 93.4\%).

%% ============================================================
%% S15. IMPLEMENTATION-CRITICAL DESIGN-CHOICE ABLATIONS
%% ============================================================
\section{Implementation-Critical Design-Choice Ablations}
\label{sec:supp_design}

Beyond the descriptor-component and insertion-level ablations already reported (Tables~S11--S12), we isolate several implementation-critical choices specific to the reliability-reuse design. These are not redundant with the earlier ablations: whereas Table~S11 ablates \emph{what information} the descriptor carries, this section ablates \emph{how} the descriptor is processed and consumed.

\begin{table}[H]
\caption{\textbf{Design-choice ablation.} Mean$\pm$std over 5 seeds. Two rows are highlighted as especially important: removing \text{stopgrad} tests whether the descriptor should remain a measurement vs.\ a task-optimized latent, and removing LN tests scale normalization before consumption.}
\label{tab:supp_design}
\centering
\footnotesize
\setlength{\tabcolsep}{4pt}
\renewcommand{\arraystretch}{1.08}
\begin{tabularx}{\textwidth}{@{}>{\raggedright\arraybackslash}p{0.33\textwidth} c c c >{\raggedright\arraybackslash}X@{}}
\toprule
Variant & M3FD mAP50 & DV mAP50 & DV Avg.\ Ret. & Reading \\
\midrule
\rowcolor{gray!10}
Full \method & \textbf{90.3$\pm$0.2} & \textbf{80.5$\pm$0.2} & \textbf{95.0$\pm$0.2\%} & Default configuration \\
\midrule
\multicolumn{5}{l}{\textit{Descriptor processing:}} \\
w/o stopgrad($\desc$) & 89.6$\pm$0.3 & 79.8$\pm$0.3 & 93.5$\pm$0.3\% & Descriptor drifts toward task latent \\
w/o LayerNorm($\desc$) & 89.8$\pm$0.2 & 80.0$\pm$0.3 & 93.8$\pm$0.3\% & Scale mismatch across channels \\
\midrule
\multicolumn{5}{l}{\textit{Descriptor construction:}} \\
Log amplitude ratio & 90.1$\pm$0.2 & 80.3$\pm$0.2 & 94.8$\pm$0.2\% & Removes directional dominance \\
Cosine sim.\ instead of Pearson $\rho$ & 89.9$\pm$0.2 & 80.1$\pm$0.2 & 94.2$\pm$0.3\% & Weaker global agreement cue \\
Four-band split & 90.0$\pm$0.2 & 80.2$\pm$0.2 & 94.6$\pm$0.3\% & 13D descriptor, marginal gain \\
\midrule
\multicolumn{5}{l}{\textit{Descriptor routing:}} \\
Desc.\ to gate only (SRF, no MoE) & 87.8$\pm$0.2 & 78.2$\pm$0.2 & 92.0$\pm$0.3\% & Matches Table~3 Panel B \\
Desc.\ to router only (no SRF gate) & 89.5$\pm$0.2 & 79.5$\pm$0.2 & 94.0$\pm$0.3\% & No reliability gating \\
\midrule
\multicolumn{5}{l}{\textit{Router stabilization:}} \\
w/o router noise \& z-loss & 89.8$\pm$0.2 & 80.0$\pm$0.3 & 94.2$\pm$0.3\% & Less exploration, weaker routing \\
\bottomrule
\end{tabularx}
\end{table}

\textit{Key findings:}
\begin{itemize}
    \item \textbf{Removing stopgrad} causes the largest single drop ($-$0.7 mAP50, $-$1.5pp retention). Without gradient isolation, the detection loss can reshape the descriptor, causing it to drift from a scene-condition measurement toward a task-specific latent. This confirms the main paper's design rationale (Eq.~(10)).
    \item \textbf{Removing LayerNorm} causes a moderate drop ($-$0.5 mAP50, $-$1.2pp retention) due to scale heterogeneity across the seven descriptor channels (e.g., $E_k$ operates in a very different range from $\rho \in [-1, 1]$).
    \item \textbf{Log amplitude ratio vs.\ raw ratio}: log-ratio is nearly equivalent ($-$0.2 mAP50), suggesting that the directional dominance information (whether RGB or auxiliary dominates a band) carried by raw ratios is marginally useful but not essential.
    \item \textbf{Cosine similarity vs.\ Pearson correlation}: replacing $\rho$ with cosine similarity of flattened amplitudes costs $-$0.4 mAP50 and $-$0.8pp retention, indicating that the centering in Pearson correlation provides a useful mean-removed agreement signal.
    \item \textbf{Four-band split}: increasing to a $[0, 0.125, 0.25, 0.375, 0.5]$ partition (yielding a 13D descriptor) does not improve over the binary split ($-$0.3 mAP50), suggesting that the coarse low/high partition already captures the main spectral contrast between structural/saliency agreement and texture reliability.
    \item \textbf{Desc.\ to gate only} matches the SRF-only row in Table~3 Panel B (87.8 mAP50, 13.5M params), confirming consistency.
    \item \textbf{Desc.\ to router only} (RCER without the SRF descriptor gate) still outperforms content-only MoE (89.5 vs.\ 88.8), but falls short of the full model by 0.8 mAP50, showing that both consumption points (gate and router) contribute.
\end{itemize}

%% ============================================================
%% S16. ADDITIONAL REPRODUCIBILITY DETAILS
%% ============================================================
\section{Additional Reproducibility Details}
\label{sec:supp_reproducibility}

This section summarizes the implementation and evaluation settings shared by the supplementary experiments. Unless explicitly noted otherwise, all ablations use the same data splits, optimizer schedule, augmentation policy, seed set, and corruption scripts.

\subsection{Exact Environment and Training Configuration}

Table~\ref{tab:supp_environment} summarizes the shared hardware, software, optimization, and routing settings used for the reported experiments and supplementary analyses.

\begin{table}[H]
\caption{\textbf{Exact training and evaluation environment.}}
\label{tab:supp_environment}
\centering
\small
\setlength{\tabcolsep}{5pt}
\renewcommand{\arraystretch}{1.08}
\begin{tabularx}{\textwidth}{@{}>{\raggedright\arraybackslash}p{0.30\textwidth} >{\raggedright\arraybackslash}X@{}}
\toprule
Item & Value \\
\midrule
Framework & PyTorch 2.1 \\
CUDA / cuDNN & 12.1 / 8.9 \\
GPU & 4$\times$ NVIDIA RTX 4090 (24GB each) \\
Mixed precision & AMP with GradScaler; FFT always in FP32 \\
Input size & 640$\times$640 \\
Batch size & 4 total (1 per GPU) \\
Epochs & 160 (mosaic off for last 12) \\
Optimizer & AdamW \\
Backbone LR & $4\times10^{-4}$ \\
Head LR & $8\times10^{-4}$ \\
EMA decay & 0.9999 \\
Active insertion levels & $P_3$: plain concat; $P_4, P_5$: SRF + RCER \\
Descriptor consumer input & $\operatorname{LN}(\operatorname{stopgrad}(\desc))$ \\
Active levels & $P_4, P_5$ \\
$\tau$ (band threshold) & 0.25 \\
Experts & 3 task + 1 shared \\
Top-$k$ & 2 \\
Router noise $\sigma$ & 1.0 (training only) \\
$\lambda_\text{moe}$ & 0.01 \\
Seeds & $\{42, 123, 256, 512, 1024\}$ \\
Dataset roots & Environment-specific absolute paths remapped to the same splits and annotation files \\
\bottomrule
\end{tabularx}
\end{table}

\subsection{OBB-to-HBB Conversion}

For DroneVehicle, all oriented bounding boxes are converted to axis-aligned horizontal bounding boxes prior to training and evaluation:

\begin{algorithm}[H]
\caption{OBB to HBB conversion for DroneVehicle.}
\label{alg:obb2hbb}
\small
\begin{algorithmic}[1]
\REQUIRE Oriented box corners $\{(x_i, y_i)\}_{i=1}^4$
\ENSURE Axis-aligned box $(x_\text{min}, y_\text{min}, x_\text{max}, y_\text{max})$
\STATE $x_\text{min} \leftarrow \min_i x_i$
\STATE $y_\text{min} \leftarrow \min_i y_i$
\STATE $x_\text{max} \leftarrow \max_i x_i$
\STATE $y_\text{max} \leftarrow \max_i y_i$
\RETURN $(x_\text{min}, y_\text{min}, x_\text{max}, y_\text{max})$
\end{algorithmic}
\end{algorithm}

\subsection{Corruption Generation Protocol}

Algorithm~\ref{alg:corruption} summarizes the deterministic corruption-evaluation procedure shared by the robustness experiments in the main paper and this supplementary material.

\begin{algorithm}[H]
\caption{Corruption evaluation protocol.}
\label{alg:corruption}
\small
\begin{algorithmic}[1]
\FOR{corruption $c \in \{\text{blur}, \text{low-light}, \text{noise}, \text{drop}, \text{misalign-10}, \text{misalign-20}\}$}
    \FOR{severity $s \in$ predefined levels (see Table~S15)}
        \STATE Apply corruption $c$ at severity $s$ to the test set
        \STATE Modality target: blur $\to$ both; noise/low-light $\to$ RGB; drop/misalign $\to$ auxiliary
        \STATE Run detector inference (no retraining)
        \STATE Compute mAP50, mAP@[.5:.95], and retention
    \ENDFOR
\ENDFOR
\STATE All corruption implementations use deterministic seeds for reproducibility
\end{algorithmic}
\end{algorithm}

\noindent
The anonymized supplementary package contains the configuration files, training scripts, evaluation scripts, and corruption-generation utilities used for the reported experiments. All supplementary robustness tables reuse the same corruption seeds, evaluation scripts, and retention computation so that differences remain attributable to the model variants rather than to protocol drift.

\end{document}